\documentclass[10pt,twocolumn,letterpaper]{article}

\usepackage[pagenumbers]{wacv} % To force page numbers, e.g. for an arXiv version

\usepackage{tabularx}
\usepackage{booktabs}
\usepackage{caption}

\usepackage{array}
\usepackage{ragged2e}

\definecolor{wacvblue}{rgb}{0.21,0.49,0.74}
\usepackage[pagebackref,breaklinks,colorlinks,allcolors=wacvblue]{hyperref}
\def\wacvPaperID{12} % *** Enter the WACV Paper ID here
\def\confName{WACV}
\def\confYear{2026}

\usepackage{times}
\usepackage{epsfig}
\usepackage{graphicx}
\usepackage{amsmath}
\usepackage{amssymb}
\usepackage{multirow}
\usepackage{bm}
\usepackage[bottom]{footmisc}
\usepackage{adjustbox}
\usepackage[normalem]{ulem}
\usepackage{subcaption}
\usepackage{parselines}
\usepackage{lineno}
\usepackage{setspace}
\usepackage{emptypage}
\usepackage{rotating}
\usepackage{makecell}
\usepackage{algorithm}
\usepackage{csquotes}
\usepackage{xhfill}
\usepackage{xcolor}
\usepackage{soul}
\usepackage{bbm}
\usepackage{dsfont}
\usepackage{float}
\usepackage{algpseudocode}
\usepackage{colortbl}
\usepackage[table,xcdraw]{xcolor} 
\usepackage{cuted} 
\usepackage{capt-of} % Often helpful for captions outside floats
\newcolumntype{Y}{>{\centering\arraybackslash}X}

\title{Agentic AI in Remote Sensing: Foundations, Taxonomy, and Emerging Systems}

\author{Niloufar Alipour Talemi \quad Julia Boone \quad Fatemeh Afghah\\
Clemson University\\
{\tt\small  \{nalipou, jcboone, fafghah\}@clemson.edu}
}

\begin{document}
\maketitle

\begin{strip}
    \centering
    % Adjust width as needed, usually 0.99\linewidth or \textwidth
    \includegraphics[width=1\linewidth]{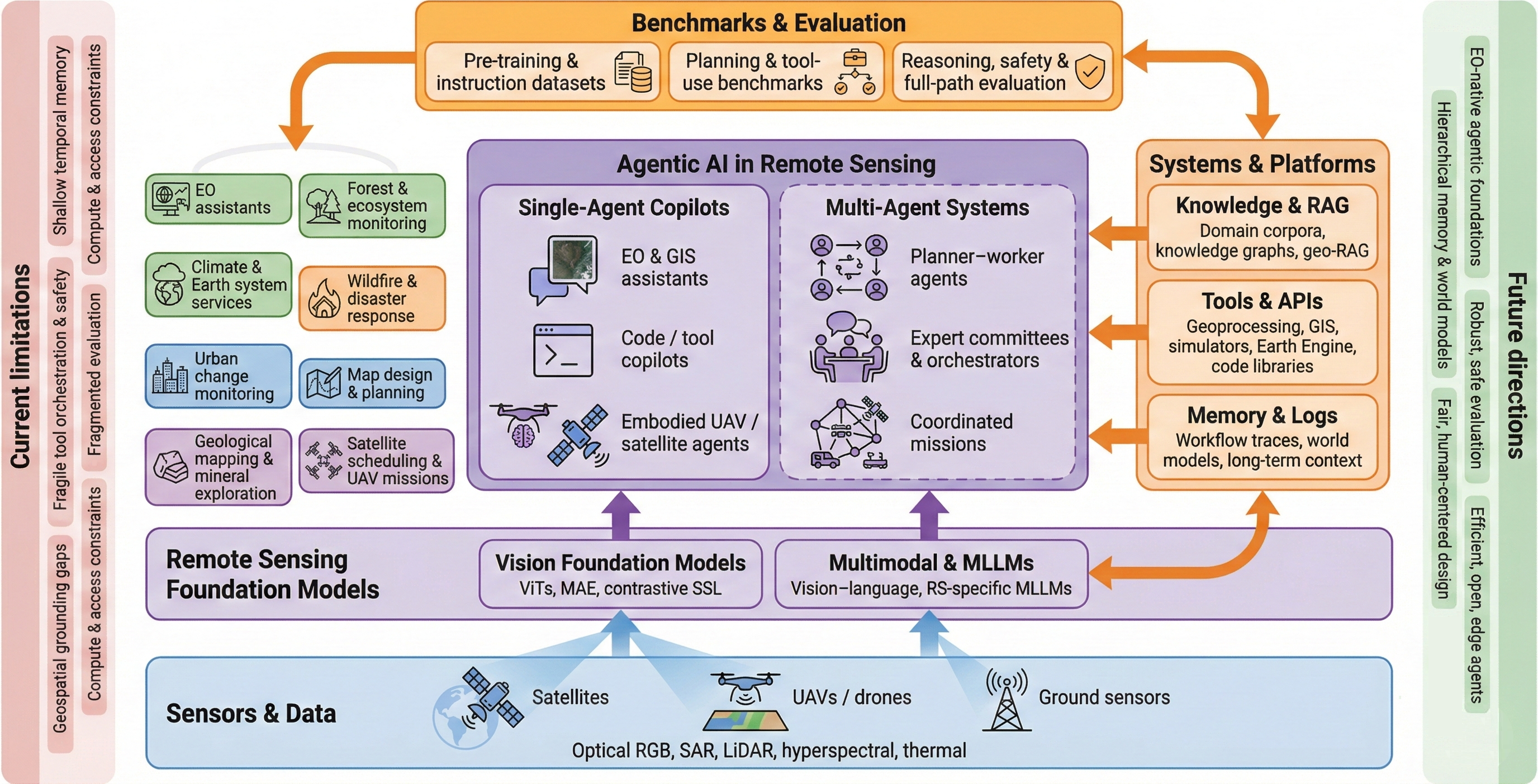} 
    \captionof{figure}{Overview of the Agentic AI ecosystem in remote sensing. The proposed framework consists of four key components: 1) Foundations: Data acquisition and foundation models; 2) Agents: A classification of systems into single-agent copilots vs. multi-agent orchestrators; 3) Systems: The technological stack (RAG, Tools, Memory) empowering the agents; and 4) Evaluation: Benchmarks for assessing planning and reasoning capabilities. The figure also maps these components to specific Earth observation applications.}
    \label{fig:over}
\end{strip}

\begin{abstract} The paradigm of Earth Observation analysis is shifting from static deep learning models to autonomous agentic AI. Although recent vision foundation models and multimodal large language models advance representation learning, they often lack the sequential planning and active tool orchestration required for complex geospatial workflows. This survey presents the first comprehensive review of agentic AI in remote sensing. We introduce a unified taxonomy distinguishing between single-agent copilots and multi-agent systems while analyzing architectural foundations such as planning mechanisms, retrieval-augmented generation, and memory structures. Furthermore, we review emerging benchmarks that move the evaluation from pixel-level accuracy to trajectory-aware reasoning correctness. By critically examining limitations in grounding, safety, and orchestration, this work outlines a strategic roadmap for the development of robust, autonomous geospatial intelligence.
\end{abstract}
\vspace{-4mm}
\section{Introduction}
\label{sec:intro}
Earth observation (EO) technologies have generated massive multi-modal remote sensing (RS) archives \cite{zhao2022overview}, ranging from very high resolution (VHR) optical imagery to synthetic-aperture radar (SAR) \cite{moreira2013tutorial}, infrared \cite{norton1991infrared}, and hyperspectral data \cite{van2012multi}. These data streams underpin critical applications in environmental monitoring \cite{pekel2016high}, disaster management \cite{ma2024transfer}, and resource exploration, making automated analysis essential. Deep learning models are the primary tools for interpreting such data, widely applied to scene classification \cite{zhao2015dirichlet}, anomaly detection \cite{wang2025non, kashiani2025roads}, change detection \cite{daudt2018urban}, and localization \cite{klemmer2025satclip}. As the field scales, there is a shift toward vision foundation models (VFMs) trained on diverse datasets to learn general-purpose representations \cite{diao2025ringmo, cong2022satmae}, with representative examples including SimCLR \cite{chen2020simple} and masked autoencoders (MAE) \cite{he2022masked}. Vision transformers (ViTs) \cite{dosovitskiy2020image} facilitate this by applying self-attention to image patches \cite{vaswani2017attention}, a capability successfully adapted to RS tasks \cite{reed2023scale, wanyan2024extending, mendieta2023towards}. Despite this progress, standard foundation models exhibit limitations. Analyses indicate that MAE-style RS models often prioritize low-level textures over global spatial structure \cite{kong2023understanding, xie2023revealing}, reducing robustness under distribution shifts \cite{park2023self}. Furthermore, many existing models rely heavily on annotated data and task-specific fine-tuning. To mitigate this annotation dependence, vision-language models (VLMs) such as CLIP \cite{radford2021learning} align image and text encoders to enable open-vocabulary detection and zero-shot segmentation \cite{liu2024remoteclip, singha2023applenet, chen2025rsrefseg}. Extending these capabilities, multimodal large language models (MLLMs) \cite{alayrac2022flamingo, wang2024qwen2, huang2023language} couple visual encoders with large language models (LLMs) \cite{bai2023qwen, brown2020language} to support complex reasoning. However, general-purpose models often degrade when applied directly to RS data due to differences in sensors, viewing geometries, and semantics.

To bridge this domain gap, recent research has developed RS-specific MLLMs such as GeoChat \cite{kuckreja2024geochat}, LHRS-Bot \cite{muhtar2024lhrs}, RS-LLaVA \cite{bazi2024rs}, SkySenseGPT \cite{luo2024skysensegpt}, and RingMoGPT \cite{hu2025ringmo}. These systems adapt architectures like LLaVA \cite{liu2023visual} or BLIP-2 \cite{li2023blip}, often using low-rank adaptation \cite{hu2022lora} to fine-tune on geospatial instructions for captioning and visual question answering (VQA). Yet, these models remain static; while they answer single-turn queries, they lack the native capacity for long-horizon memory, sequential planning, or dynamic interaction with geospatial libraries. Consequently, they fall short of handling the multi-step workflows that involve retrieval, preprocessing, and analysis and that characterize real-world geospatial operations. This limitation highlights the need to transition from static MLLMs to agentic systems.

AI agents are autonomous entities that perceive inputs, reason about tasks, and plan actions to achieve goals. In LLM-centric architectures, an agent combines a planner, a tool interface, and memory within a perception-reasoning-action loop, a paradigm that has transformed workflows in healthcare \cite{al2025agentic} and operating systems \cite{mei2024aios, achiam2023gpt}. In geospatial AI, systems such as RS-Agent \cite{xu2024rs}, GeoAgent \cite{huang2024geoagent}, Change-Agent \cite{liu2024change}, and MapBot \cite{weiss2025mapbot} instantiate this by employing LLM controllers to interpret queries and orchestrate tools for classification, segmentation, and map editing. Beyond single-agent copilots, the field is expanding toward multi-agent orchestration and realistic environments. Notable developments include GeoLLM-Engine \cite{singh2024geollm}, the multi-agent GeoLLM-Squad \cite{stamoulis2025geo}, and specialized pipelines such as RingMo-Agent \cite{hu2025ringmo} and MineAgent \cite{yu2024mineagent}. While existing surveys focus on RS foundation models and MLLMs, they typically overlook the autonomous capabilities of RS agentic systems. To the best of our knowledge, this work is the first survey dedicated to agentic AI in RS, providing a taxonomy of agent types, a comparative analysis of models and applications, and a system-level view of tools, retrieval-augmented generation (RAG) pipelines, memory mechanisms, datasets, and benchmarks. Moreover, we connect these components to emerging evaluation protocols for planning, and safety, and we articulate open challenges and future directions in geospatial grounding, long-horizon memory, and trustworthy RS agents.

\section{Background}
\label{sec:Background}
\subsection{Sensors, Data, and Applications}
RS observations are acquired by heterogeneous sensors deployed on satellites, crewed aircraft, UAVs or drones, and ground-based systems, including optical RGB and infrared cameras, multispectral and hyperspectral images, thermal sensors, LiDAR, and SAR/InSAR instruments \cite{wang2024introduction,cheng2025applications}. Each sensor type captures complementary aspects of the Earth’s surface, such as geometry from LiDAR, backscatter from SAR, or reflectance signatures from multispectral and hyperspectral instruments, yielding distinct noise characteristics, spatial resolutions, and radiometric properties that strongly influence model design and fusion strategies \cite{samadzadegan2025critical}. Platform differences in altitude, viewing geometry, coverage, and revisit time induce trade-offs between spatial resolution, temporal frequency, and swath width that must be considered when defining realistic benchmarks and when designing foundation models that can generalize across orbital, aerial, and ground perspectives \cite{wang2024aoi,lyu2022unmanned}. The combination of diverse sensors and platforms produces inherently multi-modal data, including optical and infrared images, LiDAR point clouds, hyperspectral and multispectral cubes, thermal imagery, SAR/InSAR products, and associated textual metadata or annotations \cite{hong2021overview,ma2025comprehensive,data2024multimodal}. Each modality captures distinct geophysical processes; joint exploitation significantly improves robustness and disambiguates challenging scenes, such as cloud-covered optical imagery that remains visible in SAR. These multi-modal representations support a various downstream tasks, including land-cover classification, segmentation, object detection, and change detection, as well as higher-level applications such as disaster management, urban planning, environmental monitoring, and RS question answering.

\subsection{Vision and Multimodal Foundation Models}

Foundation models \cite{bommasani2021opportunities,awais2025foundation,fei2022towards, talemi2025style, alipour2025disa} are large neural networks pre-trained on broad, heterogeneous data with generic objectives and then adapted to many downstream tasks, forming the backbone of modern vision, language, and multimodal systems. Convolutional networks such as AlexNet \cite{krizhevsky2012imagenet}, VGG \cite{simonyan2014very}, and ResNet \cite{he2016deep} pre-trained on ImageNet \cite{russakovsky2015imagenet} established transfer learning as a standard paradigm. Transformer-based VFMs such as ViT \cite{dosovitskiy2020image}, built on the transformer \cite{vaswani2017attention}, and self-supervised methods such as SimCLR \cite{chen2020simple}, MAE \cite{he2022masked}, and SimMIM \cite{xie2022simmim} enable scalable pretraining and learn transferable representations without dense labels, which is crucial for label-scarce EO and RS settings. Language foundations such as GPT-family autoregressive LLMs \cite{brown2020language,bai2023qwen} extend these ideas to web-scale text through next-token prediction and instruction tuning. Multimodal foundation models jointly learn from images and text: contrastive vision–language models such as CLIP \cite{radford2021learning} train paired image and text encoders, while MLLMs such as Flamingo \cite{alayrac2022flamingo} and LLaVA \cite{liu2023visual} couple a visual encoder with an LLM operating on unified visual–textual tokens for captioning and VQA. These mechanisms for representation learning, image–text alignment, and instruction following underpin RS-specific encoders such as SatMAE \cite{cong2022satmae} and RingMo \cite{hu2025ringmo} and geospatial MLLMs.

\section{Remote Sensing Foundation Models}

The paradigm shift initiated by large-scale pre-trained models in natural language processing and computer vision has rapidly extended to the RS domain. RS foundation models target key characteristics of geospatial data, including high-dimensional multispectral and hyperspectral signals, heterogeneous modalities such as optical imagery, SAR and LiDAR, and limited labels. Building on general VFMs such as ViT \cite{dosovitskiy2020image} and VLMs like CLIP \cite{radford2021learning}, the RS community has converged on two main adaptation strategies. The first adapts VFMs, including the segment anything model \cite{kirillov2023segment}, with self-supervised learning (SSL) techniques such as masked image modeling (MIM) \cite{hondru2025masked} and contrastive learning \cite{hu2024comprehensive} on large unlabeled RS corpora. The second designs multimodal integration frameworks that fuse RS imagery with auxiliary data, notably natural language, to construct vision-language and MLLMs for geospatial reasoning.

\subsection{Vision Foundations in Remote Sensing}
In RS, VFMs must handle modalities beyond RGB, including multispectral and hyperspectral imagery, SAR, thermal data, and LiDAR point clouds. Most RS models still initialize from ImageNet \cite{russakovsky2015imagenet}, despite its clear domain gap in modality, viewpoint, and spatial structure. RS-specific pretraining reduces this gap: MillionAID \cite{wang2022empirical} improves over ImageNet, and Satlas \cite{bastani2023satlaspretrain} shows unified multitask pretraining yields consistent gains. Because RS and EO data are abundant but labels are scarce and costly, SSL and and MIM has become the primary strategy. RS-oriented contrastive methods such as Seasonal Contrast \cite{manas2021seasonal}, Geography-Aware SSL \cite{ayush2021geography}, and SatMAE-CL \cite{cong2022satmae} learn platform-invariant, geometry-aware features across temporal and cross-sensor views. MIM approaches such as SatMAE \cite{cong2022satmae} and RingMo \cite{hu2025ringmo} reconstruct masked spatial or spectral content essential for multispectral and hyperspectral data. Collectively, these SSL paradigms drive modern RS representation learning and improve generalization across downstream tasks.

\subsection{Multi-Modal Foundations in Remote Sensing}
MLLMs extend VLMs by feeding images and other inputs, such as video and point clouds, into a language model backbone that processes a unified token sequence \cite{radford2021learning}. Visual encoders project images into the language space for joint multimodal reasoning and dialogue \cite{achiam2023gpt}, supporting tasks such as VQA and instruction following \cite{chen2023shikra}. MLLMs inherit strong reasoning ability, flexible I/O formats, and language coverage \cite{han2023imagebind}, enabling open-world interaction \cite{moon2024anymal}, multimodal assistants \cite{yuan2024osprey}, and geospatial dialogue systems \cite{kuckreja2024geochat}. In RS, MLLMs adapt this architecture to satellite and aerial imagery by combining an EO-specialized vision encoder, an alignment module, and a language model for captioning, VQA, and scene understanding \cite{pang2024h2rsvlm}. This template underlies H2RSVLM \cite{pang2024h2rsvlm}, SkyEyeGPT \cite{luo2024skysensegpt}, RSGPT \cite{hu2025rsgpt}, and EarthGPT \cite{zhang2024earthgpt}. Systems such as GeoChat \cite{kuckreja2024geochat}, LHRS Bot \cite{muhtar2024lhrs}, and RS LLaVA \cite{bazi2024rs} pair ViT or Swin \cite{liu2021Swin} encoders with alignment modules and open-source LLMs such as LLaMA \cite{touvron2023llama}, trained on instruction-tuned RS corpora \cite{bazi2024rs}. Collectively, these RS MLLMs unify classification, captioning, VQA, and grounding, advancing toward general-purpose geospatial assistants.

%%%%%%%%%%%%%%%%%%%%%%%%%%%%%%%%%%%%%%%%%%%%%%%%%%%%%%%%%%%%%%%%%%%%%%%%%%%%%%%%%%%%%%%%%%%%%%%%%%%%%%%%%%%%%%%%%%%%%%%%%%%%%

%%%%%%%%%%%%%%%%%%%%%%%%%%%%%%%%%%%%%%%%%%%%%%%%%%%%%%%%%%%%%%%%%%%%%%%%%%%%%%%%%%%%%%%%%%%%%%%%%%%%%%%%%%%%%%%%%%%%%%%%%%%%%

\section{Taxonomy of AI Agents in Remote Sensing}
Agentic AI in RS spans a broad spectrum of architectures, levels of autonomy, and application domains. In this section, we examine how foundation models are embedded in agents that perceive, reason, act, and interact with users and tools. We consider two categories: single-agent systems, where one agent plans and executes a workflow, and multi-agent systems, where multiple agents coordinate. 

\subsection{Agent Types in Remote Sensing}
\noindent\textbf{Single-agent systems.} Single-agent systems use a single controller to interpret user intent, plan analysis or control steps, call tools or models, and synthesize outputs through a unified interface without explicit collaboration between autonomous agents. RS-Agent \cite{xu2024rs} is a copilot whose LLM controller parses language queries, selects workflows from a Solution Space, and routes calls to 18 tools for enhancement, SAR detection, damage assessment, and RS-specific VQA; its DualRAG Knowledge Space grounds analysis and explanation in RS documentation, enabling task coverage without retraining \cite{xu2024rs}. Remote Sensing ChatGPT \cite{guo2024remote} adopts a similar pattern, with ChatGPT orchestrating visual models for classification, segmentation, detection, captioning, and edge extraction, and composing them via templates and descriptions into multi-step pipelines. Domain assistants such as TREE-GPT \cite{du2023tree} and Geode \cite{gupta2024geode} specialize this design. TREE-GPT \cite{du2023tree} targets forest RS by coupling an LLM with a forestry knowledge base, segmentation and LiDAR tools, and code execution for tree segmentation and ecological analysis. Geode \cite{gupta2024geode} addresses geospatial QA by compiling queries into Python programs that call “geospatial experts” over a GeoPatch abstraction and return textual answers and map visualizations. A single planning loop coordinates perception, tool use, and explanation. Embodiment and mission-level control follow this pattern in agentic UAV frameworks \cite{koubaa2025agentic}, which use an LLM-based reasoning layer to plan high-level interventions, while lower layers handle perception, integration, and control for a single UAV platform; the UAV and control stack form an embodied agent that links sensing (RGB, thermal, LiDAR) with mission-level decision-making and adapts plans as new observations arrive \cite{koubaa2025agentic}. Foundation and navigation models further support these systems. RemoteCLIP \cite{liu2024remoteclip} provides open-vocabulary, text-aligned embeddings for RS imagery, enabling text queries, novel-category localization, and encoder reuse across tasks. RingMo-Agent \cite{hu2025ringmo} unifies multi-modal encoders (optical, SAR, IR), a DeepSeek-based LLM \cite{guo2025deepseek}, and a trajectory decoder that outputs 3D waypoints for navigation and action tasks \cite{hu2025ringmo}. Although RemoteCLIP and RingMo-Agent are models rather than agents, they are frequently embedded as perception and navigation backbones inside single-agent controllers, enabling more generalizable visual and spatial reasoning.

\noindent\textbf{Multi-agent systems.} Multi-agent systems consist of autonomous agents with distinct roles that explicitly communicate and coordinate. They are ideal for decomposed workflows, heterogeneous expertise, and coordinated control of multiple assets. A common model is the planner–worker architecture. ShapefileGPT \cite{lin2025shapefilegpt} employs a planner agent to parse natural-language GIS requests, decompose them into subtasks, and oversee a worker executing Shapefile operations via a closed API. GeoJSON Agents \cite{zhang2024geojson} also divides planning and execution: a planner creates multi-step plans over GeoJSON data, while a worker performs function calls or generates code, supporting backend comparisons. Other systems further specialize roles to emulate expert teams and platform orchestrators. CartoAgent \cite{wang2025cartoagent} assigns style analysis, style-sheet and icon design, and map evaluation to separate agents. WALMAS \cite{vahidnia2025multi} forms a committee of expert agents to propose and negotiate criterion weights via Kendall’s coefficient of concordance. At platform scale, GeoFlow \cite{bhattaram2025geoflow} uses a meta-agent to build workflow graphs and dispatch subagents for data access, vision, geoprocessing, or explanation. DA4DTE \cite{tsokanaridou2025da4dte} routes queries to satellite-analysis engines (knowledge-graph, retrieval, VQA) through agents for task interpretation, routing, argument extraction, and tool feasibility. EarthLink \cite{guo2025earthlink} coordinates planning, code generation, diagnostics, and summarization, storing successful climate workflows in a reusable script library. Multi-agent formulations also appear in scientific interpretation and mission-level control. STA-CoT \cite{yu2025sta} coordinates planner, executor, and verifier agents that decompose geological questions across images, apply tools with rationales, and refine steps through targeted verification. MineAgent \cite{yu2024mineagent} combines judging agents that score mineral prospectivity from different RS and geological views with a decision agent that aggregates their semi-structured judgments, making disagreement and uncertainty explicit. Embodied controllers often use multi-agent reinforcement learning \cite{zheng2025deep}, modeling each satellite in an EO constellation as an agent with decentralized policies and a central critic for joint observation, computation, and downlink decisions \cite{zheng2025deep,dalin2020online,sautenkov2025uav}.
In UAV-CodeAgents \cite{sautenkov2025uav}, an airspace-management agent decomposes surveillance or fire-monitoring instructions into subtasks for UAV agents that execute waypoints, sense, and report, while foundation-model-backed perception and navigation are shared modules. RingMo-Agent \cite{hu2025ringmo} uses a unified encoder and trajectory decoder for optical, SAR, and infrared imagery, generating 3D waypoints while planning remains distributed.

\begin{table*}[t]
\centering
\scriptsize
\renewcommand{\arraystretch}{1.0}
\setlength{\tabcolsep}{1pt}
\resizebox{\textwidth}{!}{%
  \begin{minipage}{\textwidth}

    % header with independent column widths
       \begin{tabular}{%
     |>{\centering\arraybackslash}p{1.5cm}
      >{\centering\arraybackslash}p{1.6cm}
      >{\centering\arraybackslash}p{2.9cm}
      >{\centering\arraybackslash}p{6.1cm}
      >{\centering\arraybackslash}p{5.0cm}|}
\hline 
    \textbf{Method} &
    \textbf{Taxonomy} &
    \textbf{Applications} &
    \textbf{Systems and Technologies} &
    \textbf{Datasets and Benchmarks} \\
   \hline \hline
    \end{tabular}

    % body with original column sizes
    \begin{tabularx}{\textwidth}{%
      |>{\centering\arraybackslash}p{2cm}
      |>{\centering\arraybackslash}p{0.7cm}
      |>{\centering\arraybackslash}p{4cm}
      |>{\centering\arraybackslash}X
      |>{\centering\arraybackslash}X|}
RS-Agent \cite{xu2024rs} & SA & EO Assistants (multi-task) & RAG (DualRAG over tools and knowledge) & Classification, Detection, VQA Datasets \\
\hline
RS ChatGPT \cite{guo2024remote} & SA & EO Assistants (interactive RS dialog) & Monolithic tool ecosystems & Scene Classification, Segmentation, Detection Datasets  \\
\hline
GeoAgent \cite{chu2024geoagent} & SA & EO Assistants (GIS code reasoning) & Doc-based GIS API, Example Retrieval) & Geospatial Planning and Tool-Use \\
\hline
Geode \cite{gupta2024geode} & SA  & EO Assistants (zero-shot geospatial QA) & Expert Tools for Spatio-Temporal Retrieval) & Geospatial Planning and Tool-Use Benchmarks \\
\hline
GIS Copilot \cite{akinboyewa2025gis} & SA  & EO Assistants (GIS workflow automation) & RAG (tool and documentation grounding) & Geospatial Planning and Tool-Use Benchmarks (scripted GIS workflow) \\
\hline
TREE-GPT \cite{du2023tree} & SA & Forest and Ecosystem Monitoring & Semantic and geospatial knowledge bases (forest ontology and expert tools) & Pre-training and Instruction Tuning Datasets (forest UAV and LiDAR) \\
\hline
Earth-Agent \cite{earthbench} & SA  & EO Assistants (multi-modal analysis) & Monolithic tool ecosystems & Earth-Bench (expert-curated EO tasks, RGB, spectral, and product images)\\
\hline
Earth AI \cite{bell2025earth} & MA  & Wildfire and Disaster Monitoring & Domain-specialized planners & Reasoning Benchmarks \\
\hline
Change-Agent \cite{liu2024change} & SA  & Urban Change Monitoring & Domain-specialized planners (MCI change model coupled with LLM reasoning) & LEVIR-MCI (bi-temporal masks and captions for building and road changes) \\
\hline
DA4DTE \cite{tsokanaridou2025da4dte} & MA  & EO Assistants (digital-twin EO analysis) & Semantic and geospatial knowledge bases (geospatial knowledge graph and simulators) & Pre-training and Instruction Tuning Datasets (digital-twin scenario) \\
\hline
EarthLink \cite{guo2025earthlink} & MA & EO Assistants & Memory and Long-Term & Reasoning (long-horizon EO) \\
\hline
CartoAgent \cite{wang2025cartoagent} & MA & Map Design and Planning (Cartographic Styling) & Domain-specialized planners (style, icon, and critic agents for maps) & Reasoning Benchmarks \\
\hline
WALMAS \cite{vahidnia2025multi} & MA & Map Design and Planning & Domain-specialized planners (committee of agents with MCDA negotiation) & Geospatial Planning and Tool-Use Benchmarks \\
\hline
STA-CoT \cite{yu2025sta} & MA & Geological Mapping and Mineral Exploration & Domain-specialized planners (planner-executor-verifier with structured CoT) & MineBench (multi-image mineral exploration benchmark)\\
\hline
MineAgent \cite{yu2024mineagent} & MA & Geological Mapping and Mineral Exploration & Domain-specialized planners (hierarchical judging, decision aggregation) & MineBench (multi-image mineral exploration benchmark) \\
\hline
Wildfire Agents \cite{chen2025empowering} & SA  & Wildfire and Disaster Monitoring & RAG (LLM + geospatial wildfire and ABM knowledge) & Reasoning Benchmarks (satellite fire detections and wildfire corpora) \\
\hline
RingMo-Agent \cite{hu2025ringmo} & SA  & Satellite Scheduling and UAV Missions & Domain-specialized planners (multi-modal encoder with instruction-following policy) & RS-VL3M (3M multi-modal RS image–text pairs)\\
\hline
Agentic UAV frameworks \cite{koubaa2025agentic} & SA  & Satellite Scheduling and UAV Missions (UAV search and monitoring) & Domain-specialized planners (LLM reasoning over perception-control stack) & Geospatial Planning and Tool-Use Benchmarks (UAV mission and search scenarios) \\
\hline
GeoLLM-Engine \cite{singh2024geollm} & MA & EO Assistants (geospatial planning, tool-use environment) & Domain-specialized planners (meta-agent with workflow graphs) & GeoLLM-Engine task environment \\
\hline
GeoCode-GPT \cite{hou2025geocode} & SA  & EO Assistants (geospatial code generation and debugging) & RAG (GIS API documentation and exemplar retrieval) & GeoCode benchmark \\
\hline
GeoGraphRAG \cite{geographrag}& SA & EO Assistants 
(Geospatial modeling and code generation) & Graph-based RAG & Benchmark for geospatial modeling (300 Earth Engine workflows \\
\hline
ShapefileGPT \cite{lin2025shapefilegpt}& MA & GIS Agent (Shapefile processing and spatial analysis) & GIS tool library, internal task memory & Shapefile task dataset \\
\hline
% \midrule
% GeoBenchX \cite{1krechetova2025geobenchx} & benchmark & EO Assistants (multi-step GIS reasoning) & Graph and topology-aware retrieval & GeoBenchX tasks \\
% \hline
% GTChain \cite{zhang2025geospatial} & planner model & EO Assistants (tool-chain planning for EO workflows) & Domain-specialized planners (instruction-tuned LLM over tool-use chains) & GTChain-Eval \\
% \midrule
% GeoGraphRAG \cite{geographrag}& retrieval framework & EO Assistants (graph-based RS workflow assistance) & Graph and topology-aware retrieval (geospatial modeling knowledge graph) & Earth Engine workflow corpus \\
% \midrule
GeoLLM-Squad \cite{GeoLLM-Squad} & MA & EO Assistants & Memory and Long-Term Coherence & Reasoning Benchmarks (GeoLLM-Squad tasks and AgentSense logs) \\
% \midrule
% RS-VL3M / RemoteCLIP & datasets & EO Assistants (VL pretraining for RS tasks) & Semantic and geospatial knowledge bases (vision-language foundation encoders) & Earth-Bench \\
% ThinkGeo & benchmarks & Earth Observation Assistants (EO reasoning and tool-use evaluation) & Domain-specialized planners (ReAct-style agent trajectories and scoring) & Earth-Bench, ThinkGeo \\
% \midrule
% RescueADI / ChangeGPT bench & benchmarks & Wildfire and Disaster Monitoring (disaster and urban-change reasoning) & RAG (multi-view RS VQA and planning tasks) & RescueADI, ChangeGPT benchmark \\
% \midrule
% ShapefileGPT / GeoJSON Agents & benchmarks & Map Design and Planning (vector GIS and topology operations) & Graph and topology-aware retrieval & ShapefileGPT, GeoJSON Agents \\
% \midrule
% CORE / ToolEmu & eval frameworks & EO Assistants (tool-use safety and robustness evaluation) & Memory and Long-Term (full-path trajectory and tool-emulation checks) & CORE evaluation benchmark \\
\hline
\end{tabularx}

  \end{minipage}%
}
\caption{Summary of remote sensing agents, covering taxonomy, applications, system design, and associated datasets and benchmarks (SA/MA denote single/multi-agent).}
\label{tab:agentic_rs_methods}
\end{table*}

\subsection{Agentic AI Applications in Remote Sensing}
This subsection organizes existing systems by the concrete RS applications they target. We highlight how agentic architectures, ranging from digital copilots to embodied controllers, address specific domain challenges by coupling foundation models with specialized tools (see Table \ref{tab:agentic_rs_methods}).

\noindent\textbf{Earth Observation Assistants.} 
General-purpose intelligent assistants democratize access to RS by translating natural language into executable analyses. Systems like RS-Agent \cite{xu2024rs} and Remote Sensing ChatGPT \cite{guo2024remote} act as copilots, orchestrating tools for classification and detection without manual model selection. Similarly, GIS frameworks \cite{lin2025shapefilegpt, gupta2024geode} allow users to query vector and raster data, automating spatial joins and map generation. By bridging technical gaps, these agents facilitate rapid data retrieval and analysis, transforming static archives into interactive, dialogue-driven knowledge bases.

\noindent\textbf{Forest and Ecosystem Monitoring.} 
In forestry and ecosystem monitoring, agentic systems automate labor-intensive inventories and structural assessments. TREE-GPT \cite{du2023tree} exemplifies this by integrating vision tools with ecological knowledge bases to analyze UAV imagery and LiDAR point clouds. Beyond pixel segmentation, the agent handles tree crown delineation, biomass estimation, and health reporting via dialogue, letting foresters request stand level statistics and ecological insights.

\noindent\textbf{Climate and Earth System Services.} Agents are increasingly deployed to manage complex climate science and monitoring workflows. Systems like Earth AI \cite{bell2025earth} and Earth-Agent \cite{earthbench} introduce agentic controllers that decompose hazard questions into operations over foundation models and Earth Engine tools, automating index computation and statistical analysis. Meanwhile, digital-twin platforms such as DA4DTE \cite{tsokanaridou2025da4dte} and EarthLink \cite{guo2025earthlink} act as assistants, routing queries, planning CMIP6 experiments, and coordinating resources for disaster forecasting and environmental impact assessment with multi-source climate intelligence.

\noindent\textbf{Wildfire and Disaster Monitoring.} In the critical domain of disaster response, agentic systems connect perception to operational decisions. For wildfire management, specialized agents \cite{chen2025empowering, nie2025knowledge} go beyond hotspot detection to simulate fire spread and recommend resource allocation by fusing satellite detections with weather and infrastructure data. Similarly, in post-disaster scenarios, agents designed for adaptive interpretation \cite{liu2025rescueadi} plan rescue paths, assess damage, and turn static hazard maps into dynamic plans for time-sensitive resource allocation.

\noindent\textbf{Urban Change Monitoring.} Urban change monitoring requires agents that reason about infrastructure development rather than just pixel differences. Agents like Change-Agent \cite{liu2024change} interpret queries on urban sprawl and building updates to dynamically select segmentation or counting tools. By replacing fixed model chains with query-driven logic, they provide planners quantitative reports and semantic explanations of land-use shifts for transparent analysis beyond binary change maps.

\noindent\textbf{Map Design and Planning.} Beyond analysis, multi-agent systems are reshaping cartographic design and participatory planning. Frameworks such as CartoAgent \cite{wang2025cartoagent} employ separate agents to handle distinct design stages, from style analysis to icon generation, ensuring RS products are visualized with aesthetic and geographic precision. In spatial decision-making, collaborative agent committees \cite{vahidnia2025multi} simulate stakeholder views to negotiate criteria weights for suitability mapping.  These systems automate subjective and deliberative planning tasks, yielding consistent maps and consensus decisions for geospatial communication.

\noindent\textbf{Geological Mapping and Mineral Exploration.} In mineral exploration, agents support complex scientific reasoning over heterogeneous data. Systems like STA-CoT \cite{yu2025sta} and MineAgent \cite{yu2024mineagent} emulate geologist workflows, orchestrating analyses of structures and hyperspectral signatures across multiple images. Rather than black-box predictions, they use chain-of-thought reasoning and cross-image verification to localize deposits. By producing interpretable arguments and evidence-based recommendations, they enhance RS-driven discovery in data-scarce settings.

\noindent\textbf{Satellite Scheduling and UAV Missions.} Agentic AI is reshaping mission-level operations for satellite constellations and UAV fleets by merging perception with autonomous control. For satellites, multi-agent reinforcement learning enables cooperative scheduling of observations and downlinks under strict constraints \cite{zheng2025deep, dalin2020online}. Similarly, UAV agents \cite{koubaa2025agentic, sautenkov2025uav} utilize VLMs to decompose high-level instructions into executable flight paths. With foundation models generating continuous trajectories \cite{hu2025ringmo}, these embodied agents optimize data collection in dynamic settings where pre-planned commands fall short.

\section{Systems, Technologies, and Platforms}
\label{sec:systems}
Agentic AI in RS depends not only on multimodal models and agent architectures, but also on systems that organize knowledge, expose tools, and preserve long-horizon coherence. This section considers platform-level stacks that integrate foundation models, geospatial databases, and tool APIs, focusing on three key layers: knowledge representation and retrieval, tool/API integration, and memory for long-term coherent behavior. 

\vspace{-1mm}
\subsection{Knowledge Representation and RAG}
\noindent\textbf{Semantic and geospatial knowledge bases.}
Agentic RS systems rely on structured knowledge that grounds LLM reasoning in domain facts and geospatial context. Domain-specific bases span forestry corpora in TREE-GPT \cite{du2023tree}, which guide analysis of UAV imagery and LiDAR point clouds, RS documentation and model descriptions retrieved via DualRAG for RS-Agent and GeoAgent \cite{xu2024rs,chu2024geoagent}, and wildfire science literature for rule synthesis in simulations \cite{nie2025knowledge}. Digital twin platforms like DA4DTE expose satellite metadata as knowledge graphs, enabling agents to translate natural language into SPARQL queries over sensor, and orbit attributes. Geo-alignment promote geo-knowledge graphs encoding norms, regulations, and semantic priors as alignment targets for future geo-agents \cite{tsokanaridou2025da4dte,janowicz2025whose}.

\noindent\textbf{Retrieval-augmented generation.}
RAG has become a central mechanism for linking LLM agents to heterogeneous RS information. Earth AI combines geospatial foundation models with Gemini’s reasoning to analyze RS and population data, helping users overlay risk and vulnerability with environmental and climate forecasts, while RS-Agent, GeoAgent, and related code agents use RAG to pull technical documents, task exemplars, and executable scripts into prompts for tool selection and robust execution \cite{earthai,xu2024rs,geoagent}. Knowledge-guided wildfire modeling retrieves fire and ABM literature to derive propagation rules aligned with simulators and real events, showing how RS corpora, tool manuals, and models serve as retrievable knowledge rather than static data \cite{nie2025knowledge}.

\noindent\textbf{Graph and topology-aware retrieval.}
GeoGraphRAG \cite{geographrag} introduces a graph-based RAG pipeline where nodes are geospatial entities and edges encode spatial or functional links, enabling retrieval via graph neighborhoods beyond semantic similarity. ThinkGeo and GeoBenchX construct task graphs linking tools, images, and queries, guiding agents to retrieve over spatial, temporal, and workflow graphs for multi-hop geospatial reasoning \cite{shabbir2025thinkgeo,1krechetova2025geobenchx}.

\subsection{Tool Integration and API Orchestration}
\noindent\textbf{Monolithic tool ecosystems.} Agentic RS platforms rely on tool ecosystems that expose geoscience functions to LLM planners. Remote Sensing ChatGPT offers a toolbox of classification, segmentation, detection, captioning, edge extraction, and counting models, with ChatGPT acting as planner ~\cite{guo2024remote}. RS-Agent organizes RS tools via workflow templates in an expert-designed Solution Space \cite{xu2024rs}. Earth-Agent integrates more than one hundred geoscience tools for index computation, physical inversion, spatiotemporal statistics, and perception under a unified controller \cite{earthbench}, and digital-twin assistants such as DA4DTE route queries to engines for knowledge-graph search, retrieval, and VQA~\cite{tsokanaridou2025da4dte}. These platforms show agentic RS behavior depends on tool-layer breadth and composability.

\noindent\textbf{Domain-specialized planners.} A complementary line focuses on planner–executor interfaces in RS agents. ShapefileGPT \cite{lin2025shapefilegpt} and GeoJSON \cite{luo2025geojson} exemplify patterns where a planner parses user intent and a worker calls GIS APIs or emits geospatial code. GTChain \cite{zhang2025geospatial} instruction-tunes an open LLM on tool-use chains to output ordered tool sequences that surpass larger closed models. GeoFlow \cite{bhattaram2025geoflow} models tasks as workflow graphs with nodes defining subagents, APIs, and parameters for function calls. GIS Copilot \cite{akinboyewa2025gis} synthesizes and debugs PyQGIS scripts with validators that enforce coordinate and topology rules, showing how tool schemas and robust execution enable RS pipelines. 

\noindent\textbf{Memory and Long-Term Coherence.} Memory in agentic RS systems spans task-level context, graph-structured workflow knowledge, and platform-level logs; Agents log intermediate results, tool calls, and plans, as seen in RS-Agent’s Solution Space and DualRAG Knowledge Space \cite{xu2024rs}, ChangeGPT’s urban-change dialog logs \cite{xiao2025llm}, STA-CoT and MineAgent’s scored spatial tuples with rationales \cite{yu2025sta,yu2024mineagent}, and EarthLink’s script archive \cite{guo2025earthlink}. At platform scale, Earth-Bench, GeoCode, GeoBenchX, and ThinkGeo record tool-call trajectories as workflows \cite{earthbench,hou2025geocode,1krechetova2025geobenchx,shabbir2025thinkgeo}, GTChain \cite{zhang2025geospatial} treats tool-use chains as offline memory of geospatial processing \cite{zhang2025geospatial}, and wildfire-response agents store daily fire descriptors and analog events for multi-day decisions \cite{chen2025empowering}. Graph-based systems like GeoGraphRAG \cite{geographrag} encode expert scripts in a geospatial modeling knowledge graph, while multi-agent platforms like GeoLLM-Squad \cite{GeoLLM-Squad} and participatory urban sensing systems such as AgentSense \cite{guo2025agentsense} maintain workflow or meta-operation memories for retrieval-guided orchestration and adaptation, supporting coherence, reproducibility, and auditing.

\begin{figure}
    \centering
    \includegraphics[width=0.92\linewidth]{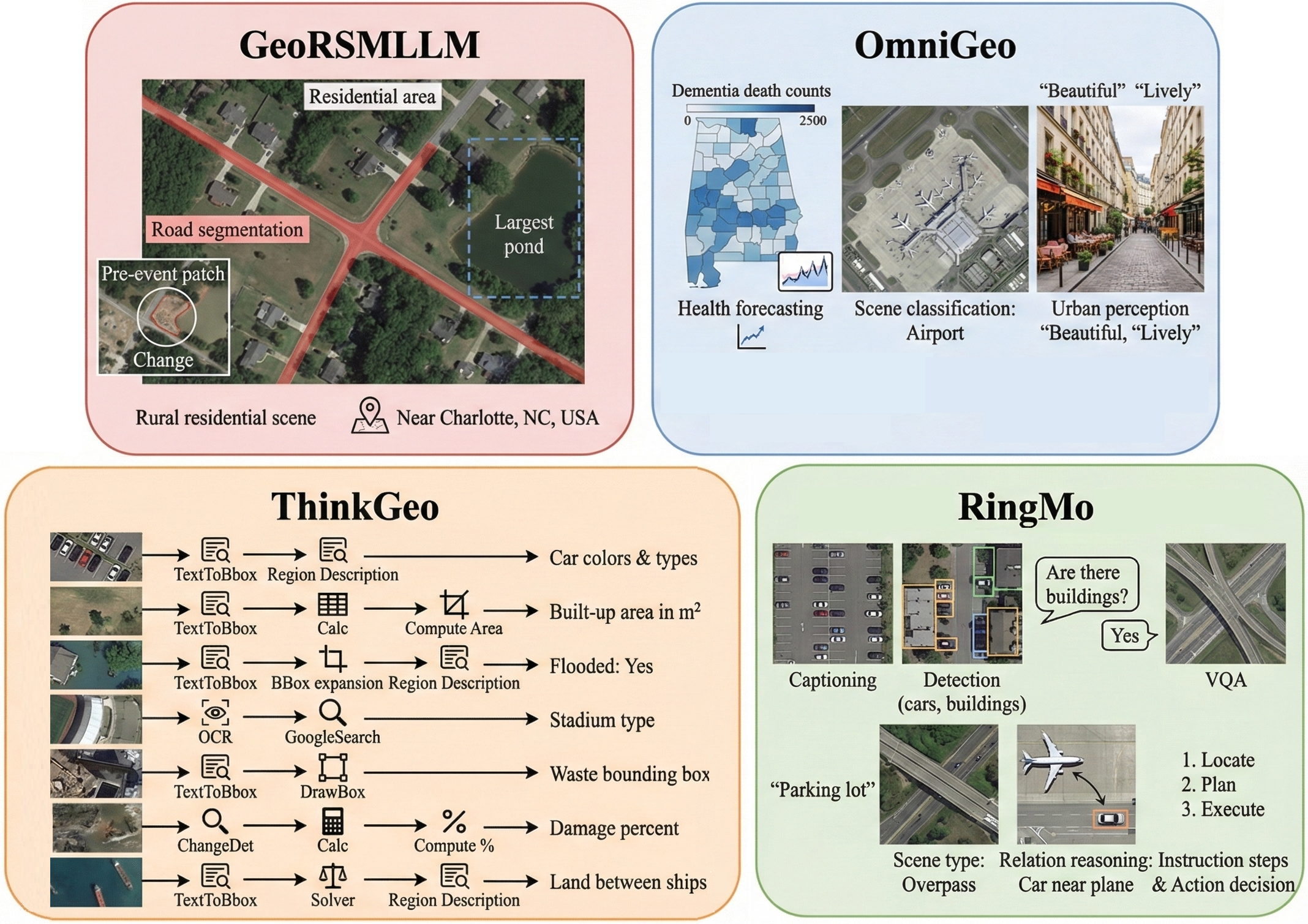}
    \caption{Benchmarks and datasets for agentic remote sensing AI. GeoRSMLLM \cite{zhang2025georsmllm} includes referring-expression tasks, change detection, scene classification, and geo-localization; OmniGeo \cite{yuan2025omnigeo} covers health geography, RS scene classification, urban perception, and geospatial semantics; ThinkGeo \cite{shabbir2025thinkgeo} pairs RS patches with multi-tool reasoning; and RingMo-Agent \cite{hu2025ringmo} supports multi-modal RS tasks such as relation reasoning.}
    \label{fig:placeholder}
\end{figure}

\section{Benchmarks and Evaluations}
\subsection{Datasets and Benchmarks}
The shift from static perception models to autonomous agents requires evaluation of planning, tool use, and reasoning over pixel accuracy or single-turn captioning and VQA. This subsection reviews datasets and benchmarks for evaluating such agentic capabilities in EO and RS (see Fig. \ref{fig:placeholder}).

\noindent\textbf{Pre-training and Instruction Tuning Datasets.} Pre-training and instruction-tuning datasets align visual semantics with language instructions and provide the basis for agentic reasoning. RS-VL3M \cite{hu2025ringmo} aggregates millions of optical, SAR, and infrared image–text pairs for diverse tasks, while RemoteCLIP \cite{liu2024remoteclip} addresses RS data scarcity via a mask-to-caption pipeline that turns segmentation masks into text for contrastive vision–language pre-training. Multimodal datasets from GeoRSMLLM \cite{zhang2025georsmllm}, LHRS-Bot-Nova \cite{li2025lhrs}, and OmniGeo \cite{yuan2025omnigeo} further supply large-scale instruction-tuning data for RS MLLMs.

\noindent\textbf{Geospatial Planning and Tool-Use Benchmarks.} Geospatial planning and tool-use benchmarks test agents' ability to build and run geospatial workflows. GeoLLM-Engine \cite{singh2024geollm} offers a large task environment with a model-checker for verifying final states, while GeoCode \cite{chu2024geoagent} assesses execution-based synthesis across 19,000 tasks and 28 libraries. GeoBenchX \cite{1krechetova2025geobenchx} measures multi-step reasoning and epistemic awareness using unsolvable queries, and GTChain-Eval \cite{zhang2025geospatial} scores tool-chain logic. Additional evaluations include GeoTool-GPT’s benchmark \cite{wei2025geotool} and GeoGraphRAG’s Earth Engine workflows \cite{geographrag}, grounded in a geospatial modeling knowledge graph.

\noindent\textbf{Reasoning Benchmarks.} Reasoning benchmarks evaluate domain-specific multi-step inference and process quality. Earth-Bench \cite{earthbench} and ThinkGeo \cite{shabbir2025thinkgeo} score RGB and SAR tasks using ReAct-style tool use. RS MLLM benchmarks, including grounding datasets from GeoChat \cite{kuckreja2024geochat} and the LHRS-Bench suite in LHRS-Bot-Nova \cite{li2025lhrs}, test region-level reasoning, spatial relations, and instruction following \cite{kuckreja2024geochat,li2025lhrs}. RescueADI \cite{liu2025rescueadi} and ChangeGPT \cite{xiao2025llm} address hazard and urban-change analysis, while MineBench \cite{yu2024mineagent,yu2025sta} evaluates geological and hyperspectral reasoning. Vector-focused benchmarks such as ShapefileGPT \cite{lin2025shapefilegpt} and GeoJSON Agents \cite{zhang2024geojson} examine precise geometric operations under function-calling and code-generation settings. Frameworks such as CORE and ToolEmu \cite{CORE,toolemu} add safety-focused evaluation via full-path correctness and harmful-call detection. System-level studies like GeoLLM-Squad \cite{GeoLLM-Squad}, AgentSense \cite{guo2025agentsense}, and smart-city platforms \cite{xu2025agentic} report correctness, coverage, and planner-aligned performance which are critical for RS agents.

\subsection{Evaluation}
Evaluating agentic RS systems requires moving from static assessments to trajectory-aware protocols that validate internal reasoning. Unlike traditional benchmarks focused only on output correctness \cite{Lobry2020RSVQA, xu2024rs}, safety-critical workflows need full-path evaluation for reliability.

\noindent\textbf{Evaluation Paradigms: Full-Path versus End-to-End.} Unlike traditional end-to-end metrics focused on final outputs, agentic systems require full-path evaluation of reasoning and safety. CORE formalizes this using Deterministic Finite Automata to detect forbidden transitions via valid state graphs \cite{CORE}. Bridging these, Earth-Agent and ThinkGeo employ dual-level scoring and LLM-as-a-Judge methods to verify both procedural integrity and semantic correctness for reliable, efficient operation \cite{earthbench, shabbir2025thinkgeo}.

\noindent\textbf{Key Metrics for Agentic Correctness and Robustness.} Evaluating reasoning performance requires metrics that distinguish between planning and execution errors. Benchmarks increasingly adopt step-by-step metrics like Tool Accuracy, which correlates strongly with final answer accuracy, and Argument Accuracy, which identifies syntactic errors in function parameters \cite{shabbir2025thinkgeo, huang2024geoagent}. For code-generating agents, standard metrics include Pass@k and Task Completion Rate, supplemented by analyses of failure types such as API hallucinations and stagnation loops \cite{GeoAgent-Auto}.

\noindent\textbf{Traditional Remote Sensing Metrics and Alignment.}
Trajectory metrics must be paired with RS measures. Perception components are evaluated using F1-score, mAP@0.5, accuracy, and relative error in tasks such as counting and biomass estimation \cite{xu2024rs, hu2025ringmo}. System-level evaluations measure how agentic mistakes affect geophysical products. GeoLLM-Squad introduces a mean-square percentage error to quantify error propagation into land surface temperature, and tree loss \cite{GeoLLM-Squad}.

\section{Limitations and Future Directions}

\subsection{Limitations of Current Agentic RS Systems}
\noindent\textbf{Limited Geospatial Grounding.}
RS specific systems such as RS-Agent and GeoAgent operate largely on RGB imagery and vector data, with limited support for spectral products, SAR and multi sensor stacks \cite{xu2024rs,geoagent}. Earth-Agent reports degradation on non RGB products and quantitative queries, showing that current backbones and tool prompts fail to cover EO diversity \cite{earthbench}.

\noindent\textbf{Fragile Tool Orchestration.}
Tool use in current RS agents is fragile. ThinkGeo \cite{shabbir2025thinkgeo}, GeoBenchX \cite{1krechetova2025geobenchx}, GeoCode \cite{hou2025geocode} and GTChain-Eval \cite{zhang2025geospatial} report errors in tool selection and argument formatting. CORE \cite{CORE} and ToolEmu \cite{toolemu} show that agents often ignore preconditions, repeat failing calls or trigger inappropriate tools, directly affecting satellite tasking, UAV routes and access to sensitive data.

\noindent\textbf{Shallow Temporal Memory.} Most RS agents maintain only short context and flat logs of tool calls, so Earth-Bench \cite{earthbench} and ThinkGeo \cite{shabbir2025thinkgeo} report repeated downloads, redundant computation and inconsistent reuse of intermediate results. Multi temporal stacks and mission context are seldom stored in structured, queryable memory, undermining robustness in wildfire management and change analysis.

\noindent\textbf{Fragmented Evaluation Protocols.} GeoLLM-Engine \cite{singh2024geollm}, GeoBenchX \cite{1krechetova2025geobenchx} and Earth-Bench \cite{earthbench} cover disjoint parts of the design space and still lack a unified protocol that measures planning quality, perception accuracy and safety across RS missions. Many evaluations report only final answers, often on synthetic scenarios, leaving robustness to data drift and adversarial inputs in use unknown.

\noindent\textbf{Compute Constraints.}
Many agentic pipelines depend on large proprietary LLMs and cloud infrastructures, which impose latency and resource constraints, while open source alternatives lag on challenging planning and multi tool code generation \cite{1krechetova2025geobenchx,earthai,zhang2025geospatial}.

\subsection{Future Directions for Agentic RS}
\noindent\textbf{Foundations and Memory.}
A key direction is to build EO native foundation models instead of adapting natural image systems. Earth-Agent already couples RGB and spectral encoders with a structured tool ecosystem and trajectory aware evaluation, outlining an integrated stack where perception, tool use and validation are jointly designed \cite{earthbench}. Extending such models with multi sensor encoders over optical, SAR and thermal data, together with physics informed surrogates and simulators, would let agents connect raw measurements, derived products and scientific reasoning within one workflow. Open, research oriented counterparts of platforms such as Google Earth AI are also needed for transparent experimentation and shared benchmarks \cite{earthai}. Long running missions further require hierarchical memory that blends vector stores, geo knowledge graphs and workflows. 

\noindent\textbf{Safety, Efficiency, Equity.}
As deployment progresses, future work should treat robustness, safety and efficiency as central design goals for agentic RS systems. Benchmark suites such as ThinkGeo, GeoBenchX and Earth-Bench should be extended with diverse tasks, harder negative examples and targeted stress tests for distribution shift, adversarial prompts and unsolvable queries, plus standardised reporting of trajectory metrics, harmful call rates and resource consumption to support safety-critical certification \cite{shabbir2025thinkgeo,1krechetova2025geobenchx}. To bridge cloud and edge deployments, agents will need smaller language models and distilled planners that preserve reliable tool use on constrained hardware, with frameworks such as GeoCode-GPT and GTChain providing templates for locally deployable geospatial agents \cite{1krechetova2025geobenchx,hou2025geocode,zhang2025geospatial}.

\section{Conclusion} Agentic AI marks a pivotal evolution in RS, advancing from static perception to autonomous, goal-directed decision-making. This survey has reviewed this emerging landscape by defining a taxonomy of single-agent copilots and multi-agent systems while analyzing the essential infrastructure of planning, memory, and RAG. Although current systems demonstrate impressive capabilities in code generation and analysis, they face critical challenges in geospatial grounding, safety, and long-horizon coherence. Addressing these gaps through Earth-native models and rigorous evaluation will enable transition from prototypes to trustworthy agents capable of complex planetary-scale operations.

\section*{Acknowledgment}
This material is based upon work supported by the National Science Foundation under Grant Numbers CNS-2232048, and CNS-2204445 and NASA under the award number 80NSSC23K1393. 
\newpage
\twocolumn[
  \begin{center}
    \LARGE \textbf{Supplementary Material for Agentic AI in Remote Sensing: Foundations, Taxonomy, and Emerging Systems} \\

    \vspace{20pt}
  \end{center}
]

\appendix

% Restart Table numbering and change format to Table S1, S2 etc. (optional) or just 1, 2
\setcounter{table}{0}
\renewcommand{\thetable}{\arabic{table}}

\section{RS Datasets Across Applications}
\label{sec:supp_rs_datasets}
In this section, we cover representative remote sensing (RS) datasets that ground the application taxonomy in the main paper. In Table~\ref{tab:rs_datasets_taxonomy}, we group benchmarks across scene classification, semantic segmentation, object detection, change detection, building and road extraction, disaster and hazard mapping, text–image grounding, and earth observation (EO) foundation pretraining. The table lists each dataset’s sensor modality, spatial resolution, and benchmark task. It groups together aerial RGB scene datasets \cite{yang2010bag,xia2017aid,cheng2017remote}, sentinel-based LULC collections \cite{helber2019eurosat,demir2018deepglobe}, disaster-focused resources such as xBD, FloodNet, and Sen1Floods11 \cite{gupta2019xbd,rahnemoonfar2021floodnet,bonafilia2020sen1floods11}, and large EO pretraining corpora including SSL4EO-S12 and EarthView \cite{wang2023ssl4eo,velazquez2025earthview}. Collectively, these datasets offer a practical catalog for connecting specific RS tasks with suitable sensors and benchmarks when developing and evaluating new methods.

\section{Datasets and Benchmarks for Agentic RS}
\label{sec:supp_agentic_benchmarks}
In this section, we cover datasets and evaluation suites that explicitly target LLM-driven agentic methods in geospatial and RS. In Table~\ref{tab:agentic_rs_methods}, we summarize benchmarks for geospatial tool use and multi-step reasoning, including GeoBenchX \cite{1krechetova2025geobenchx} and GTChain-IT / CTChain-Eval \cite{zhang2025geospatial}, multi-turn multimodal dialogue over SAR and infrared imagery in RS-VL3M \cite{hu2025ringmo}, and realistic tool-augmented task suites in ThinkGeo and RescueADI \cite{shabbir2025thinkgeo,liu2025rescueadi}. The table further includes ShapefileGPT for Shapefile-based spatial analysis \cite{lin2025shapefilegpt} and generic tool-use evaluation frameworks like CORE \cite{CORE}. Taken together, these benchmarks provide a focused basis for assessing agentic behavior in RS and for comparing emerging systems under consistent evaluation protocols.

%%%%%%%%%%%%%%%%%%%%

\begin{table*}[h]
\centering
\scriptsize
\setlength{\tabcolsep}{4pt}
\begin{tabularx}{\textwidth}{>{\centering\arraybackslash}p{3.8cm} >{\centering\arraybackslash}p{3.5cm} >{\centering\arraybackslash}p{3.4cm} Y}

\toprule
\textbf{Dataset} & \textbf{Sensor / Modality} & \textbf{Resolution / Scale} & \textbf{Dataset Application} \\
\midrule \midrule
\multicolumn{4}{l}{\textbf{Scene / LULC classification}} \\
UC Merced Land Use \cite{yang2010bag} & Aerial RGB & $\sim$0.3 m, 256$\times$256 patches & land-use scene classification (21 classes) \\
AID \cite{xia2017aid} & Aerial RGB & 600$\times$600 pixel patches & Aerial scene image classification (30 classes) \\
NWPU-RESISC45 \cite{cheng2017remote} & Aerial RGB & 256$\times$256 pixel patches & Scene classification (45 classes) \\
EuroSAT \cite{helber2019eurosat} & Sentinel-2 multispectral & $64 \times 64$ pixel patches & Land Use and Land Cover (LULC) classification (10 classes) \\
Million-AID \cite{long2021creating} & Aerial RGB & Variable (0.5m to 153m) & Eerial scene classification (51 classes) \\
MLRSNet \cite{qi2020mlrsnet} & Optical Satellite/ Aerial RGB & 256$\times$256 pixel patches & Multi-label semantic scene understanding (46 scene categories, 60 labels)\\
\midrule
\multicolumn{4}{l}{\textbf{Semantic segmentation (urban, LULC)}} \\
Inria Aerial Image Labeling \cite{maggiori2017can}& Aerial RGB & 5000$\times$5000 px, 0.3 m/pixel & Semantic segmentation \\
DeepGlobe Land Cover \cite{demir2018deepglobe} & Satellite RGB & 2448$\times$2448 px, 0.5 m/pixel & Rural land cover semantic segmentation \\
LoveDA \cite{wang2021loveda}& Spaceborne RGB satellite & 1024$\times$1024 px, 0.3 m/pixel & Land-cover segmentation under domain shift (rural/urban) \\

DynamicEarthNet \cite{toker2022dynamicearthnet} & Planet multi-spectral satellite & 1024$\times$1024 px, 3 m GSD & LULC semantic and change segmentation. \\
Dynamic World \cite{brown2022dynamic}& Sentinel-2 multi-spectral images & Global 10 m/pixel & Near real-time LULC mapping \\
\midrule
\multicolumn{4}{l}{\textbf{Object detection / instance segmentation}} \\

DOTA \cite{xia2018dota}& Optical aerial/satellite imagery (RGB/gray) & High-resolution, variable up to 20k & Oriented object detection in aerial images \\

xView \cite{lam2018xview} & WorldView-3 satellite imagery & 0.3 m GSD, 1 km$^2$ chips & Overhead multi-class object detection \\
FAIR1M \cite{sun2022fair1m}& High-res optical satellite & 0.3–0.8 m GSD, 1k–10k pixel & Fine-grained oriented object detection, classification \\
\midrule
\multicolumn{4}{l}{\textbf{Change detection (bi-/multi-temporal)}} \\
LEVIR-CD \cite{chen2020spatial} & Google Earth VHR RGB & 1024$\times$1024 pixel, 0.5 m/pixel & Bitemporal building change segmentation \\
SYSU-CD \cite{shi2021deeply}& 0.5 m RGB aerial imagery & 256$\times$256 pixel, 0.5 m GSD & Bitemporal high-resolution change detection \\
S2Looking \cite{shen2021s2looking} & Side-looking RGB optical satellite imagery  & 1024$\times$1024, 0.5–0.8 m GSD & Bitemporal building change detection \\
OSCD (Onera)\cite{daudt2018urban} & Sentinel-2 multispectral optical imagery & 600$\times$600 at 10m resolution & Urban binary change detection. \\
% SpaceNet-7 & WorldView multi-temporal & Multi-year data & Building footprint evolution and change analysis \\
\midrule
\multicolumn{4}{l}{\textbf{Building / road extraction}} \\
DeepGlobe \cite{demir2018deepglobe} & Satellite Optical RGB & 2448$\times$2448 pixel, 0.5 m/pixel & Rural land cover segmentation. \\
DeepGlobe Road \cite{DeepGlobe18} & Satellite RGB & 1024$\times$1024 px tiles, ~0.5 m/pixel & Road and street network extraction \\
CrowdAI \cite{adimoolam2023efficient}& RGB satellite imagery & 300$\times$300 pixel tiles, ~0.3 m GSD & Building footprint detection / segmentation \\
\midrule
\multicolumn{4}{l}{\textbf{Disaster, damage, hazard mapping}} \\
xBD \cite{gupta2019xbd} & Multispectral satellite imagery & $\leq 0.8$ m GSD & Building damage assessment, change detection \\
FloodNet \cite{rahnemoonfar2021floodnet} & UAV RGB & 4000$\times$3000 pixel, 1.5 cm GSD & Post-flood damage segmentation and VQA \\
Sen1Floods11 \cite{bonafilia2020sen1floods11} & Sentinel-1 SAR imagery & 512$\times$512 chips, 120406 km$^2$ global & Flood and permanent water segmentation \\
UrbanSARFloods \cite{zhao2024urbansarfloods} & Sentinel-1 SAR & 512$\times$512 chips, 807500 km$^2$ & Urban and open-area flood segmentation \\
FireRisk \cite{shen2023firerisk}& NAIP aerial RGB & 270$\times$270 px tiles, ~1 m & Wildfire risk level classification \\

\midrule 

\multicolumn{4}{l}{\textbf{Text–image, captioning, VQA}} \\
RSICD \cite{yamani2024remote} & Aerial / satellite RGB & Patch-level & RS image captioning and text–image alignment \\
% RSVQA-LR / RSVQA-HR & Sentinel-2 and others & Low- and high-res & Remote sensing VQA with automatically generated Q\&A \\
RSIVQA \cite{zheng2021mutual} & Multi-source aerial / satellite RGB imagery & Variable, ~0.1-8 m GSD& VQA for RS scene understanding \\
% LEVIR-VQA / LEVER-VQA & Bi-temporal optical & Change scenes & VQA on change detection and bi-temporal reasoning \\
FloodNet-VQA \cite{rahnemoonfar2021floodnet} & UAV RGB aerial & 4000$\times$3000 px, ~1.5 cm GSD & Post-flood scene understanding, segmentation, VQA \\
\midrule
\multicolumn{4}{l}{\textbf{Pretraining corpora / EO foundation}} \\
SSL4EO-S12 \cite{wang2023ssl4eo} & Sentinel-1 SAR, Sentinel-2 multispectral & 264$\times$ 264 pixel, 2640$\times$2640 m &Self-supervised EO pretraining, downstream tasks 
Elib DLR
+1 \\
EarthView \cite{velazquez2025earthview} & Multisource optical RS & Mixed 1-30 m GSD, global & Self-supervised pretraining for EO \\
% Google Satellite Embedding Dataset & Multi-source EO embeddings & Global 10 m grid & Pixel-level geospatial embeddings for downstream tasks \\
% DynamicEarthNet \cite{} & Planet + Sentinel & Regional time series & Joint pretraining/evaluation for semantic and temporal tasks \\
\midrule \bottomrule
\end{tabularx}

\caption{Representative benchmarks and datasets for remote sensing, grouped by application category (shown as section headers). The table highlights typical sensors, spatial scale, and primary benchmark tasks to support method selection and evaluation design.}
\label{tab:rs_datasets_taxonomy}
\end{table*}

%%%%%%%%%%%%%%%%%%%%%%%%%%%%

\begin{table*}[t]
\centering
\footnotesize
\renewcommand{\arraystretch}{1.0}
\setlength{\tabcolsep}{1pt}
\resizebox{\textwidth}{!}{%
\begin{tabularx}{\textwidth}{%
  >{\centering\arraybackslash}m{2cm}
  |>{\centering\arraybackslash}m{2.3cm}
 | >{\centering\arraybackslash}m{5.0cm}
  |>{\centering\arraybackslash}m{\dimexpr\linewidth-9.5cm}%
}
\hline \hline
\multicolumn{1}{c}{} &
\textbf{Dataset / Benchmark} &
\textbf{Applications} &
\textbf{Systems and Technologies} \\
\hline
GeoBenchX \cite{1krechetova2025geobenchx} &
Dataset and evaluation framework &
Multi-step GIS reasoning &
LangGraph ReAct agent, Python geospatial stack, and an LLM as Judge \\
\hline
GTChain-IT / CTChain-Eval \cite{zhang2025geospatial} &
Dataset and evaluation framework &
Benchmarking LLMs on geospatial tool use tasks &
Simulated tool-use environment and fixed GIS tool APIs \\
\hline
RS-VL3M \cite{hu2025ringmo} &
Benchmark &
Benchmark for multi turn dialogue over SAR/IR with joint perception &
Infrared RS images with scene labels, combined with SAR-CLA and optical benchmarks in multi modality \\
\hline
ThinkGeo \cite{shabbir2025thinkgeo} &
Benchmark &
Benchmark to evaluate tool-augmented LLM agents on realistic remote sensing tasks &
ReAct tool-calling with AgentLego tools, RGB/SAR imagery \\
\hline
RescueADI \cite{liu2025rescueadi} &
Benchmarks &
Adaptive disaster interpretation &
PSPNet, GroundingDINO, counting and area tools \\
\hline
Shapefile \cite{lin2025shapefilegpt} &
Benchmark &
Benchmarking on 42 Shapefile spatial analysis tasks &
27-function Shapefile GIS tool library \\
\hline
CORE \cite{CORE} &
Eval frameworks &
Evaluation framework for tool-using agents &
Simulated tool APIs with CORE path metrics \\
\hline \hline
\end{tabularx}%
}
\caption{Overview of datasets and evaluation benchmarks for LLM driven agentic methods in geospatial and remote sensing.}
\label{tab:agentic_rs_methods}
\end{table*}

{
    \small
    \bibliographystyle{ieeenat_fullname}
    \bibliography{main}

@String(CVPR= {IEEE Conf. Comput. Vis. Pattern Recog.})

@String(ICCV= {Int. Conf. Comput. Vis.})

@String(ICLR = {Int. Conf. Learn. Represent.})

@String(AAAI = {AAAI})

@String(CVPR  = {CVPR})

@String(ICCV  = {ICCV})

@String(ICLR  = {ICLR})

@article{zhao2022overview,
  title={An overview of the applications of earth observation satellite data: impacts and future trends},
  author={Zhao, Qiang and Yu, Le and Du, Zhenrong and Peng, Dailiang and Hao, Pengyu and Zhang, Yongguang and Gong, Peng},
  journal={Remote Sensing},
  volume={14},
  number={8},
  pages={1863},
  year={2022},
  publisher={MDPI}
}

@article{moreira2013tutorial,
  title={A tutorial on synthetic aperture radar},
  author={Moreira, Alberto and Prats-Iraola, Pau and Younis, Marwan and Krieger, Gerhard and Hajnsek, Irena and Papathanassiou, Konstantinos P},
  journal={IEEE Geoscience and remote sensing magazine},
  volume={1},
  number={1},
  pages={6--43},
  year={2013},
  publisher={IEEE}
}

@article{norton1991infrared,
  title={Infrared image sensors},
  author={Norton, Paul R},
  journal={Optical Engineering},
  volume={30},
  number={11},
  pages={1649--1663},
  year={1991},
  publisher={SPIE}
}

@article{van2012multi,
  title={Multi-and hyperspectral geologic remote sensing: A review},
  author={Van der Meer, Freek D and Van der Werff, Harald MA and Van Ruitenbeek, Frank JA and Hecker, Chris A and Bakker, Wim H and Noomen, Marleen F and Van Der Meijde, Mark and Carranza, E John M and De Smeth, J Boudewijn and Woldai, Tsehaie},
  journal={International journal of applied Earth observation and geoinformation},
  volume={14},
  number={1},
  pages={112--128},
  year={2012},
  publisher={Elsevier}
}

@article{pekel2016high,
  title={High-resolution mapping of global surface water and its long-term changes},
  author={Pekel, Jean-Fran{\c{c}}ois and Cottam, Andrew and Gorelick, Noel and Belward, Alan S},
  journal={Nature},
  volume={540},
  number={7633},
  pages={418--422},
  year={2016},
  publisher={Nature Publishing Group UK London}
}

@article{ma2024transfer,
  title={Transfer learning in environmental remote sensing},
  author={Ma, Yuchi and Chen, Shuo and Ermon, Stefano and Lobell, David B},
  journal={Remote Sensing of Environment},
  volume={301},
  pages={113924},
  year={2024},
  publisher={Elsevier}
}

@article{zhao2015dirichlet,
  title={Dirichlet-derived multiple topic scene classification model for high spatial resolution remote sensing imagery},
  author={Zhao, Bei and Zhong, Yanfei and Xia, Gui-Song and Zhang, Liangpei},
  journal={IEEE Transactions on Geoscience and Remote Sensing},
  volume={54},
  number={4},
  pages={2108--2123},
  year={2015},
  publisher={IEEE}
}

@inproceedings{daudt2018urban,
  title={Urban change detection for multispectral earth observation using convolutional neural networks},
  author={Daudt, Rodrigo Caye and Le Saux, Bertr and Boulch, Alexandre and Gousseau, Yann},
  booktitle={IGARSS 2018-2018 IEEE International Geoscience and Remote Sensing Symposium},
  pages={2115--2118},
  year={2018},
  organization={Ieee}
}

@inproceedings{liu2021Swin,
  title={{Swin} Transformer: Hierarchical Vision Transformer using Shifted Windows},
  author={Liu, Ze and Lin, Yutong and Cao, Yue and Hu, Han and Wei, Yixuan and Zhang, Zheng and Lin, Stephen and Guo, Baining},
  booktitle={Proceedings of the IEEE/CVF International Conference on Computer Vision (ICCV)},
  year={2021}
}

@article{touvron2023llama,
  title={{LLaMA}: Open and efficient foundation language models},
  author={Touvron, Hugo and Lavril, Thibaut and Izacard, Gautier and Martinet, Xavier and Lachaux, Marie-Anne and Lacroix, Timoth{\'e}e and Rozi{\`e}re, Baptiste and Goyal, Naman and Hambro, Eric and Azhar, Faisal and others},
  journal={arXiv preprint arXiv:2302.13971},
  year={2023}
}

@inproceedings{klemmer2025satclip,
  title={Satclip: Global, general-purpose location embeddings with satellite imagery},
  author={Klemmer, Konstantin and Rolf, Esther and Robinson, Caleb and Mackey, Lester and Ru{\ss}wurm, Marc},
  booktitle={Proceedings of the AAAI Conference on Artificial Intelligence},
  volume={39},
  number={4},
  pages={4347--4355},
  year={2025}
}

@article{diao2025ringmo,
  title={RingMo-aerial: An aerial remote sensing foundation model with affine transformation contrastive learning},
  author={Diao, Wenhui and Yu, Haichen and Kang, Kaiyue and Ling, Tong and Liu, Di and Feng, Yingchao and Bi, Hanbo and Ren, Libo and Li, Xuexue and Mao, Yongqiang and others},
  journal={IEEE Transactions on Pattern Analysis and Machine Intelligence},
  year={2025},
  publisher={IEEE}
}

@article{cong2022satmae,
  title={{SatMAE}: Pre-training transformers for temporal and multi-spectral satellite imagery},
  author={Cong, Yezhen and Khanna, Samar and Meng, Chenlin and Liu, Patrick and Rozi, Erik and He, Yutong and Burke, Marshall and Lobell, David and Ermon, Stefano},
  journal={Advances in Neural Information Processing Systems},
  volume={35},
  pages={197--211},
  year={2022}
}

@inproceedings{
dosovitskiy2020image,
title={An Image is Worth 16x16 Words: Transformers for Image Recognition at Scale},
author={Alexey Dosovitskiy and Lucas Beyer and Alexander Kolesnikov and Dirk Weissenborn and Xiaohua Zhai and Thomas Unterthiner and Mostafa Dehghani and Matthias Minderer and Georg Heigold and Sylvain Gelly and Jakob Uszkoreit and Neil Houlsby},
booktitle={International Conference on Learning Representations},
year={2021},
url={https://openreview.net/forum?id=YicbFdNTTy}
}

@article{vaswani2017attention,
  title={Attention is all you need},
  author={Vaswani, Ashish and Shazeer, Noam and Parmar, Niki and Uszkoreit, Jakob and Jones, Llion and Gomez, Aidan N and Kaiser, {\L}ukasz and Polosukhin, Illia},
  journal={Advances in neural information processing systems},
  volume={30},
  year={2017}
}

@inproceedings{reed2023scale,
  title={{Scale-MAE}: A scale-aware masked autoencoder for multiscale geospatial representation learning},
  author={Reed, Colorado J and Gupta, Ritwik and Li, Shufan and Brockman, Sarah and Funk, Christopher and Clipp, Brian and Keutzer, Kurt and Candido, Salvatore and Uyttendaele, Matt and Darrell, Trevor},
  booktitle={Proceedings of the IEEE/CVF International Conference on Computer Vision},
  pages={4088--4099},
  year={2023}
}

@inproceedings{wanyan2024extending,
  title={Extending global-local view alignment for self-supervised learning with remote sensing imagery},
  author={Wanyan, Xinye and Seneviratne, Sachith and Shen, Shuchang and Kirley, Michael},
  booktitle={Proceedings of the IEEE/CVF Conference on Computer Vision and Pattern Recognition},
  pages={2443--2453},
  year={2024}
}

@inproceedings{chen2020simple,
  title={A simple framework for contrastive learning of visual representations},
  author={Chen, Ting and Kornblith, Simon and Norouzi, Mohammad and Hinton, Geoffrey},
  booktitle={International conference on machine learning},
  pages={1597--1607},
  year={2020},
  organization={PmLR}
}

@inproceedings{he2022masked,
  title={Masked autoencoders are scalable vision learners},
  author={He, Kaiming and Chen, Xinlei and Xie, Saining and Li, Yanghao and Doll{\'a}r, Piotr and Girshick, Ross},
  booktitle={Proceedings of the IEEE/CVF conference on computer vision and pattern recognition},
  pages={16000--16009},
  year={2022}
}

@inproceedings{mendieta2023towards,
  title={Towards geospatial foundation models via continual pretraining},
  author={Mendieta, Mat{\'\i}as and Han, Boran and Shi, Xingjian and Zhu, Yi and Chen, Chen},
  booktitle={Proceedings of the IEEE/CVF International Conference on Computer Vision},
  pages={16806--16816},
  year={2023}
}

@inproceedings{kong2023understanding,
  title={Understanding masked autoencoders via hierarchical latent variable models},
  author={Kong, Lingjing and Ma, Martin Q and Chen, Guangyi and Xing, Eric P and Chi, Yuejie and Morency, Louis-Philippe and Zhang, Kun},
  booktitle={Proceedings of the IEEE/CVF Conference on Computer Vision and Pattern Recognition},
  pages={7918--7928},
  year={2023}
}

@inproceedings{xie2023revealing,
  title={Revealing the dark secrets of masked image modeling},
  author={Xie, Zhenda and Geng, Zigang and Hu, Jingcheng and Zhang, Zheng and Hu, Han and Cao, Yue},
  booktitle={Proceedings of the IEEE/CVF conference on computer vision and pattern recognition},
  pages={14475--14485},
  year={2023}
}

@article{park2023self,
  title={What do self-supervised vision transformers learn?},
  author={Park, Namuk and Kim, Wonjae and Heo, Byeongho and Kim, Taekyung and Yun, Sangdoo},
  journal={arXiv preprint arXiv:2305.00729},
  year={2023}
}

@inproceedings{singha2023applenet,
  title={Applenet: Visual attention parameterized prompt learning for few-shot remote sensing image generalization using clip},
  author={Singha, Mainak and Jha, Ankit and Solanki, Bhupendra and Bose, Shirsha and Banerjee, Biplab},
  booktitle={Proceedings of the IEEE/CVF Conference on Computer Vision and Pattern Recognition},
  pages={2024--2034},
  year={2023}
}

@article{chen2025rsrefseg,
  title={Rsrefseg: Referring remote sensing image segmentation with foundation models},
  author={Chen, Keyan and Zhang, Jiafan and Liu, Chenyang and Zou, Zhengxia and Shi, Zhenwei},
  journal={arXiv preprint arXiv:2501.06809},
  year={2025}
}

@article{bai2023qwen,
  title={Qwen technical report},
  author={Bai, Jinze and Bai, Shuai and Chu, Yunfei and Cui, Zeyu and Dang, Kai and Deng, Xiaodong and Fan, Yang and Ge, Wenbin and Han, Yu and Huang, Fei and others},
  journal={arXiv preprint arXiv:2309.16609},
  year={2023}
}

@article{brown2020language,
  title={Language models are few-shot learners},
  author={Brown, Tom and Mann, Benjamin and Ryder, Nick and Subbiah, Melanie and Kaplan, Jared D and Dhariwal, Prafulla and Neelakantan, Arvind and Shyam, Pranav and Sastry, Girish and Askell, Amanda and others},
  journal={Advances in neural information processing systems},
  volume={33},
  pages={1877--1901},
  year={2020}
}

@article{alayrac2022flamingo,
  title={Flamingo: a visual language model for few-shot learning},
  author={Alayrac, Jean-Baptiste and Donahue, Jeff and Luc, Pauline and Miech, Antoine and Barr, Iain and Hasson, Yana and Lenc, Karel and Mensch, Arthur and Millican, Katherine and Reynolds, Malcolm and others},
  journal={Advances in neural information processing systems},
  volume={35},
  pages={23716--23736},
  year={2022}
}

@article{wang2024qwen2,
  title={Qwen2-vl: Enhancing vision-language model's perception of the world at any resolution},
  author={Wang, Peng and Bai, Shuai and Tan, Sinan and Wang, Shijie and Fan, Zhihao and Bai, Jinze and Chen, Keqin and Liu, Xuejing and Wang, Jialin and Ge, Wenbin and others},
  journal={arXiv preprint arXiv:2409.12191},
  year={2024}
}

@article{huang2023language,
  title={Language is not all you need: Aligning perception with language models},
  author={Huang, Shaohan and Dong, Li and Wang, Wenhui and Hao, Yaru and Singhal, Saksham and Ma, Shuming and Lv, Tengchao and Cui, Lei and Mohammed, Owais Khan and Patra, Barun and others},
  journal={Advances in Neural Information Processing Systems},
  volume={36},
  pages={72096--72109},
  year={2023}
}

@inproceedings{kuckreja2024geochat,
  title={Geochat: Grounded large vision-language model for remote sensing},
  author={Kuckreja, Kartik and Danish, Muhammad Sohail and Naseer, Muzammal and Das, Abhijit and Khan, Salman and Khan, Fahad Shahbaz},
  booktitle={Proceedings of the IEEE/CVF Conference on Computer Vision and Pattern Recognition},
  pages={27831--27840},
  year={2024}
}

@inproceedings{muhtar2024lhrs,
  title={{LHRS-Bot}: Empowering remote sensing with {VGI}-Enhanced large multimodal language model},
  author={Muhtar, Dilxat and Li, Zhenshi and Gu, Feng and Zhang, Xueliang and Xiao, Pengfeng},
  booktitle={European Conference on Computer Vision},
  pages={440--457},
  year={2024},
  organization={Springer}
}

@article{liu2023visual,
  title={Visual instruction tuning},
  author={Liu, Haotian and Li, Chunyuan and Wu, Qingyang and Lee, Yong Jae},
  journal={Advances in neural information processing systems},
  volume={36},
  pages={34892--34916},
  year={2023}
}

@article{russakovsky2015imagenet,
  title={{ImageNet} large scale visual recognition challenge},
  author={Russakovsky, Olga and Deng, Jia and Su, Hao and Krause, Jonathan and Satheesh, Sanjeev and Ma, Sean and Huang, Zhiheng and Karpathy, Andrej and Khosla, Aditya and Bernstein, Michael and others},
  journal={International journal of computer vision},
  volume={115},
  number={3},
  pages={211--252},
  year={2015},
  publisher={Springer}
}

@inproceedings{li2023blip,
  title={Blip-2: Bootstrapping language-image pre-training with frozen image encoders and large language models},
  author={Li, Junnan and Li, Dongxu and Savarese, Silvio and Hoi, Steven},
  booktitle={International conference on machine learning},
  pages={19730--19742},
  year={2023},
  organization={PMLR}
}

@article{hu2022lora,
  title={Lora: Low-rank adaptation of large language models.},
  author={Hu, Edward J and Shen, Yelong and Wallis, Phillip and Allen-Zhu, Zeyuan and Li, Yuanzhi and Wang, Shean and Wang, Lu and Chen, Weizhu and others},
  journal={ICLR},
  volume={1},
  number={2},
  pages={3},
  year={2022}
}

@article{al2025agentic,
  title={Agentic Large-Language-Model Systems in Medicine: A Systematic Review and Taxonomy},
  author={Al Radi, Abdul Mohaimen and Cao, Xu and Yu, Fanyang and Liu, Yuyuan and Liu, Fengbei and Wang, Chong and Chen, Yuanhong and Chen, Jintai and Wang, Hu and Meng, Yanda and others},
  journal={Authorea Preprints},
  year={2025},
  publisher={Authorea}
}

@article{mei2024aios,
  title={Aios: Llm agent operating system},
  author={Mei, Kai and Zhu, Xi and Xu, Wujiang and Hua, Wenyue and Jin, Mingyu and Li, Zelong and Xu, Shuyuan and Ye, Ruosong and Ge, Yingqiang and Zhang, Yongfeng},
  journal={arXiv preprint arXiv:2403.16971},
  year={2024}
}

@inproceedings{huang2024geoagent,
  title={{GeoAgent}: To empower llms using geospatial tools for address standardization},
  author={Huang, Chenghua and Chen, Shisong and Li, Zhixu and Qu, Jianfeng and Xiao, Yanghua and Liu, Jiaxin and Chen, Zhigang},
  booktitle={Findings of the Association for Computational Linguistics ACL 2024},
  pages={6048--6063},
  year={2024}
}

@article{liu2024change,
  title={Change-agent: Towards interactive comprehensive remote sensing change interpretation and analysis},
  author={Liu, Chenyang and Chen, Keyan and Zhang, Haotian and Qi, Zipeng and Zou, Zhengxia and Shi, Zhenwei},
  journal={IEEE Transactions on Geoscience and Remote Sensing},
  year={2024},
  publisher={IEEE}
}

@inproceedings{weiss2025mapbot,
  title={MapBot: A Multi-Modal Agent for Geospatial Analysis},
  author={Weiss, Martin and Rahaman, Nasim and Pal, Chris},
  booktitle={Proceedings of the 24th International Conference on Autonomous Agents and Multiagent Systems},
  pages={3059--3061},
  year={2025}
}

@inproceedings{stamoulis2025geo,
  title={Geo-olm: Enabling sustainable earth observation studies with cost-efficient open language models \& state-driven workflows},
  author={Stamoulis, Dimitrios and Marculescu, Diana},
  booktitle={Proceedings of the ACM SIGCAS/SIGCHI Conference on Computing and Sustainable Societies},
  pages={608--619},
  year={2025}
}

@article{samadzadegan2025critical,
  title={A critical review on multi-sensor and multi-platform remote sensing data fusion approaches: current status and prospects},
  author={Samadzadegan, Farhad and Toosi, Ahmad and Dadrass Javan, Farzaneh},
  journal={International journal of remote sensing},
  volume={46},
  number={3},
  pages={1327--1402},
  year={2025},
  publisher={Taylor \& Francis}
}

@article{wang2024aoi,
  title={AoI analysis of satellite--UAV synergy real-time remote sensing system},
  author={Wang, Libo and Zhang, Xiangyin and Qin, Kaiyu and Wang, Zhuwei and Zhou, Jiayi and Song, Deyu},
  journal={Remote Sensing},
  volume={16},
  number={17},
  pages={3305},
  year={2024},
  publisher={MDPI}
}

@article{lyu2022unmanned,
  title={Unmanned aerial vehicle (UAV) remote sensing in grassland ecosystem monitoring: A systematic review},
  author={Lyu, Xin and Li, Xiaobing and Dang, Dongliang and Dou, Huashun and Wang, Kai and Lou, Anru},
  journal={Remote Sensing},
  volume={14},
  number={5},
  pages={1096},
  year={2022},
  publisher={MDPI}
}

@inproceedings{hong2021overview,
  title={An overview of multimodal remote sensing data fusion: From image to feature, from shallow to deep},
  author={Hong, Danfeng and Chanussot, Jocelyn and Zhu, Xiao Xiang},
  booktitle={2021 IEEE International Geoscience and Remote Sensing Symposium IGARSS},
  pages={1245--1248},
  year={2021},
  organization={IEEE}
}

@article{ma2025comprehensive,
  title={A Comprehensive Review of Multi-Source Data Fusion Processing Methods},
  author={Ma, Xiaping and Zhou, Peimin and He, Xiaoxing and Zhang, Sheng},
  year={2025},
  publisher={Preprints}
}

@article{data2024multimodal,
  title={Multimodal artificial intelligence foundation models: Unleashing the power of remote sensing big data in earth observation},
  author={DATA, MRSB},
  journal={Innovation},
  volume={2},
  number={1},
  pages={100055},
  year={2024}
}

@book{wang2024introduction,
  title={Introduction to LiDAR remote sensing},
  author={Wang, Cheng and Yang, Xuebo and Xi, Xiaohuan and Nie, Sheng and Dong, Pinliang},
  year={2024},
  publisher={CRC Press Boca Raton, FL, USA}
}

@article{cheng2025applications,
  title={Applications and advancements of spaceborne InSAR in landslide monitoring and susceptibility mapping: a systematic review},
  author={Cheng, Yusen and Pang, Hongli and Li, Yangyang and Fan, Lei and Wei, Shengjie and Yuan, Ziwen and Fang, Yinqing},
  journal={Remote Sensing},
  volume={17},
  number={6},
  pages={999},
  year={2025},
  publisher={MDPI}
}

@inproceedings{kirillov2023segment,
  title={Segment anything},
  author={Kirillov, Alexander and Mintun, Eric and Ravi, Nikhila and Mao, Hanzi and Rolland, Chloe and Gustafson, Laura and Xiao, Tete and Whitehead, Spencer and Berg, Alexander C and Lo, Wan-Yen and others},
  booktitle={Proceedings of the IEEE/CVF international conference on computer vision},
  pages={4015--4026},
  year={2023}
}

@article{hondru2025masked,
  title={Masked image modeling: A survey},
  author={Hondru, Vlad and Croitoru, Florinel Alin and Minaee, Shervin and Ionescu, Radu Tudor and Sebe, Nicu},
  journal={International Journal of Computer Vision},
  volume={133},
  number={10},
  pages={7154--7200},
  year={2025},
  publisher={Springer}
}

@article{fei2022towards,
  title={Towards artificial general intelligence via a multimodal foundation model},
  author={Fei, Nanyi and Lu, Zhiwu and Gao, Yizhao and Yang, Guoxing and Huo, Yuqi and Wen, Jingyuan and Lu, Haoyu and Song, Ruihua and Gao, Xin and Xiang, Tao and others},
  journal={Nature Communications},
  volume={13},
  number={1},
  pages={3094},
  year={2022},
  publisher={Nature Publishing Group UK London}
}

@article{awais2025foundation,
  title={Foundation models defining a new era in vision: a survey and outlook},
  author={Awais, Muhammad and Naseer, Muzammal and Khan, Salman and Anwer, Rao Muhammad and Cholakkal, Hisham and Shah, Mubarak and Yang, Ming-Hsuan and Khan, Fahad Shahbaz},
  journal={IEEE Transactions on Pattern Analysis and Machine Intelligence},
  year={2025},
  publisher={IEEE}
}

@article{hu2024comprehensive,
  title={A comprehensive survey on contrastive learning},
  author={Hu, Haigen and Wang, Xiaoyuan and Zhang, Yan and Chen, Qi and Guan, Qiu},
  journal={Neurocomputing},
  volume={610},
  pages={128645},
  year={2024},
  publisher={Elsevier}
}

@article{wang2022empirical,
  title={An empirical study of remote sensing pretraining},
  author={Wang, Di and Zhang, Jing and Du, Bo and Xia, Gui-Song and Tao, Dacheng},
  journal={IEEE Transactions on Geoscience and Remote Sensing},
  volume={61},
  pages={1--20},
  year={2022},
  publisher={IEEE}
}

@inproceedings{bastani2023satlaspretrain,
  title={Satlaspretrain: A large-scale dataset for remote sensing image understanding},
  author={Bastani, Favyen and Wolters, Piper and Gupta, Ritwik and Ferdinando, Joe and Kembhavi, Aniruddha},
  booktitle={Proceedings of the IEEE/CVF International Conference on Computer Vision},
  pages={16772--16782},
  year={2023}
}

@inproceedings{manas2021seasonal,
  title={Seasonal Contrast: Unsupervised pre-training from uncurated remote sensing data},
  author={Ma{\~n}as, Oscar and Lacoste, Alexandre and Giro-i-Nieto, Xavier and Vazquez, David and Rodriguez, Pablo},
  booktitle={CVPR},
  year={2021}
}

@inproceedings{ayush2021geography,
  title={Geography-aware self-supervised learning},
  author={Ayush, Kumar and Uzkent, Burak and Meng, Chenlin and Tanmay, Shah and Burke, Marshall and Lobell, David and Ermon, Stefano},
  booktitle={ICCV},
  year={2021}
}

@inproceedings{radford2021learning,
  title={Learning transferable visual models from natural language supervision},
  author={Radford, Alec and Kim, Jong Wook and Hallacy, Chris and Ramesh, Aditya and Goh, Gabriel and Agarwal, Sandhini and Sastry, Girish and Askell, Amanda and Mishkin, Pamela and Clark, Jack and others},
  booktitle={International conference on machine learning},
  pages={8748--8763},
  year={2021},
  organization={PmLR}
}

@article{achiam2023gpt,
  title={Gpt-4 technical report},
  author={Achiam, Josh and Adler, Steven and Agarwal, Sandhini and Ahmad, Lama and Akkaya, Ilge and Aleman, Florencia Leoni and Almeida, Diogo and Altenschmidt, Janko and Altman, Sam and Anadkat, Shyamal and others},
  journal={arXiv preprint arXiv:2303.08774},
  year={2023}
}

@article{chen2023shikra,
  title={Shikra: Unleashing multimodal llm's referential dialogue magic},
  author={Chen, Keqin and Zhang, Zhao and Zeng, Weili and Zhang, Richong and Zhu, Feng and Zhao, Rui},
  journal={arXiv preprint arXiv:2306.15195},
  year={2023}
}

@article{han2023imagebind,
  title={Imagebind-llm: Multi-modality instruction tuning},
  author={Han, Jiaming and Zhang, Renrui and Shao, Wenqi and Gao, Peng and Xu, Peng and Xiao, Han and Zhang, Kaipeng and Liu, Chris and Wen, Song and Guo, Ziyu and others},
  journal={arXiv preprint arXiv:2309.03905},
  year={2023}
}

@inproceedings{moon2024anymal,
  title={Anymal: An efficient and scalable any-modality augmented language model},
  author={Moon, Seungwhan and Madotto, Andrea and Lin, Zhaojiang and Nagarajan, Tushar and Smith, Matt and Jain, Shashank and Yeh, Chun-Fu and Murugesan, Prakash and Heidari, Peyman and Liu, Yue and others},
  booktitle={Proceedings of the 2024 Conference on Empirical Methods in Natural Language Processing: Industry Track},
  pages={1314--1332},
  year={2024}
}

@inproceedings{yuan2024osprey,
  title={Osprey: Pixel understanding with visual instruction tuning},
  author={Yuan, Yuqian and Li, Wentong and Liu, Jian and Tang, Dongqi and Luo, Xinjie and Qin, Chi and Zhang, Lei and Zhu, Jianke},
  booktitle={Proceedings of the IEEE/CVF Conference on Computer Vision and Pattern Recognition},
  pages={28202--28211},
  year={2024}
}

@article{pang2024h2rsvlm,
  title={{H2RSVLM}: Towards helpful and honest remote sensing large vision language model},
  author={Pang, Chao and Wu, Jiang and Li, Jiayu and Liu, Yi and Sun, Jiaxing and Li, Weijia and Weng, Xingxing and Wang, Shuai and Feng, Litong and Xia, Gui-Song and others},
  journal={CoRR},
  year={2024}
}

@article{luo2024skysensegpt,
  title={{SkySenseGPT}: A fine-grained instruction tuning dataset and model for remote sensing vision-language understanding},
  author={Luo, Junwei and Pang, Zhen and Zhang, Yongjun and Wang, Tingzhu and Wang, Linlin and Dang, Bo and Lao, Jiangwei and Wang, Jian and Chen, Jingdong and Tan, Yihua and others},
  journal={arXiv preprint arXiv:2406.10100},
  year={2024}
}

@article{hu2025rsgpt,
  title={{RSGPT}: A remote sensing vision language model and benchmark},
  author={Hu, Yuan and Yuan, Jianlong and Wen, Congcong and Lu, Xiaonan and Liu, Yu and Li, Xiang},
  journal={ISPRS Journal of Photogrammetry and Remote Sensing},
  volume={224},
  pages={272--286},
  year={2025},
  publisher={Elsevier}
}

@article{zhang2024earthgpt,
  title={{EarthGPT}: A universal multimodal large language model for multisensor image comprehension in remote sensing domain},
  author={Zhang, Wei and Cai, Miaoxin and Zhang, Tong and Zhuang, Yin and Mao, Xuerui},
  journal={IEEE Transactions on Geoscience and Remote Sensing},
  volume={62},
  pages={1--20},
  year={2024},
  publisher={IEEE}
}

@article{bazi2024rs,
  title={{RS-LLaVA}: A large vision-language model for joint captioning and question answering in remote sensing imagery},
  author={Bazi, Yakoub and Bashmal, Laila and Al Rahhal, Mohamad Mahmoud and Ricci, Riccardo and Melgani, Farid},
  journal={Remote Sensing},
  volume={16},
  number={9},
  year={2024},
  publisher={MDPI}
}

@article{liu2025rescueadi,
  title={{RescueADI}: adaptive disaster interpretation in remote sensing images with autonomous agents},
  author={Liu, Zhuoran and Zhao, Danpei and Yuan, Bo and Jiang, Zhiguo},
  journal={IEEE Transactions on Geoscience and Remote Sensing},
  year={2025},
  publisher={IEEE}
}

@inproceedings{yu2025sta,
  title={{STA-CoT}: Structured Target-Centric Agentic Chain-of-Thought for Consistent Multi-Image Geological Reasoning},
  author={Yu, Beibei and Shen, Tao and Chen, Ling},
  booktitle={Findings of the Association for Computational Linguistics: EMNLP 2025},
  pages={25426--25444},
  year={2025}
}

@article{lin2025shapefilegpt,
  title={{ShapefileGPT}: A multi-agent large language model framework for automated shapefile processing},
  author={Lin, Qingming and Hu, Rui and Li, Huaxia and Wu, Sensen and Li, Yadong and Fang, Kai and Feng, Hailin and Du, Zhenhong and Xu, Liuchang},
  journal={International Journal of Digital Earth},
  volume={18},
  number={2},
  pages={2577884},
  year={2025},
  publisher={Taylor \& Francis}
}

@article{zhang2024geojson,
  title={GeoJSON Agents: A Multi-Agent LLM Architecture for Geospatial Analysis-Function Calling vs Code Generation},
  author={Luo, Qianqian and Xu, Liuchang and Lin, Qingming and Wu, Sensen and Mao, Ruichen and Wang, Chao and Feng, Hailin and Huang, Bo and Du, Zhenhong},
  journal={arXiv preprint arXiv:2509.08863},
  year={2025}
}

@article{yu2024mineagent,
  title={{MineAgent}: Towards Remote-Sensing Mineral Exploration with Multimodal Large Language Models},
  author={Yu, Beibei and Shen, Tao and Na, Hongbin and Chen, Ling and Li, Denqi},
  journal={arXiv preprint arXiv:2412.17339},
  year={2024},
  url={https://arxiv.org/abs/2412.17339}
}

@inproceedings{singh2024geollm,
  title={{GeoLLM-Engine}: A realistic environment for building geospatial copilots},
  author={Singh, Simranjit and Fore, Michael and Stamoulis, Dimitrios},
  booktitle={Proceedings of the IEEE/CVF Conference on Computer Vision and Pattern Recognition},
  pages={585--594},
  year={2024}
}

@article{GeoLLM-Squad,
  title={Multi-Agent Geospatial Copilots for Remote Sensing Workflows},
  author={Lee, Chaehong and Paramanayakam, Varatheepan and Karatzas, Andreas and Jian, Yanan and Foret, Michael and Liao, Heming and Yu, Fuxun and Li, Ruopu and Anagnostopoulos, Iraklis and Stamoulis, Dimitrios},
  journal={arXiv preprint arXiv:2501.16254},
  year={2025},
  url={https://arxiv.org/abs/2501.16254}
}

@article{GeoAgent-Auto,
  title={An {LLM} Agent for Automatic Geospatial Data Analysis},
  author={Chen, Yuxing and Wang, Weijie and Lobry, Sylvain and Kurtz, Camille},
  journal={arXiv preprint arXiv:2410.18792},
  year={2024},
  url={https://arxiv.org/abs/2410.18792}
}

@article{Lobry2020RSVQA,
  title={{RSVQA}: Visual question answering for remote sensing data},
  author={Lobry, Sylvain and Marcos, Diego and Murray, Jesse and Tuia, Devis},
  journal={IEEE Transactions on Geoscience and Remote Sensing},
  volume={58},
  number={12},
  pages={8555--8566},
  year={2020}
}

@article{zheng2025deep,
  title={Deep Reinforcement Learning for Joint Observation and On-Orbit Computation Scheduling in Agile Satellite Constellations},
  author={Zheng, Lujie and Jiang, Qiangqiang and Zhang, Yamin and Chen, Bo},
  journal={Aerospace},
  volume={12},
  number={10},
  pages={914},
  year={2025},
  publisher={MDPI}
}

@article{du2023tree,
  title={{TREE-GPT}: Modular large language model expert system for forest remote sensing image understanding and interactive analysis},
  author={Du, Siqi and Tang, Shengjun and Wang, Weixi and Li, Xiaoming and Guo, Renzhong},
  journal={arXiv preprint arXiv:2310.04698},
  year={2023}
}

@article{dalin2020online,
  title={An online distributed satellite cooperative observation scheduling algorithm based on multiagent deep reinforcement learning},
  author={Dalin, Li and Haijiao, Wang and Zhen, Yang and Yanfeng, Gu and Shi, Shen},
  journal={IEEE Geoscience and Remote Sensing Letters},
  volume={18},
  number={11},
  pages={1901--1905},
  year={2020},
  publisher={IEEE}
}

@article{chen2025empowering,
  title={Empowering LLM Agents with Geospatial Awareness: Toward Grounded Reasoning for Wildfire Response},
  author={Chen, Yiheng and Li, Lingyao and Ma, Zihui and Hu, Qikai and Zhu, Yilun and Deng, Min and Yu, Runlong},
  journal={arXiv preprint arXiv:2510.12061},
  year={2025}
}

@article{koubaa2025agentic,
  title={Agentic {UAVs}: LLM-Driven Autonomy with Integrated Tool-Calling and Cognitive Reasoning},
  author={Koubaa, Anis and Gabr, Khaled},
  journal={arXiv preprint arXiv:2509.13352},
  year={2025}
}

@article{sautenkov2025uav,
  title={{UAV-CodeAgents}: Scalable uav mission planning via multi-agent react and vision-language reasoning},
  author={Sautenkov, Oleg and Yaqoot, Yasheerah and Mustafa, Muhammad Ahsan and Batool, Faryal and Sam, Jeffrin and Lykov, Artem and Wen, Chih-Yung and Tsetserukou, Dzmitry},
  journal={arXiv preprint arXiv:2505.07236},
  year={2025}
}

@article{bell2025earth,
  title={Earth AI: Unlocking Geospatial Insights with Foundation Models and Cross-Modal Reasoning},
  author={Bell, Aaron and Aides, Amit and Helmy, Amr and Muslim, Arbaaz and Barzilai, Aviad and Slobodkin, Aviv and Jaber, Bolous and Schottlander, David and Leifman, George and Paul, Joydeep and others},
  journal={arXiv preprint arXiv:2510.18318},
  year={2025}
}

@article{janowicz2025whose,
  title={Whose Truth? Pluralistic {Geo-Alignment} for (Agentic) AI},
  author={Janowicz, Krzysztof and Liu, Zilong and Mai, Gengchen and Wang, Zhangyu and Majic, Ivan and Fortacz, Alexandra and McKenzie, Grant and Gao, Song},
  journal={arXiv preprint arXiv:2508.05432},
  year={2025}
}

@inproceedings{nie2025knowledge,
  title={{Knowledge-Guided} Large Language Models for Enhancing Agent-Based Wildfire Spatial Simulation},
  author={Nie, Ying and Gao, Song},
  booktitle={Proceedings of the 8th ACM SIGSPATIAL International Workshop on Geospatial Simulation},
  pages={49--56},
  year={2025}
}

@article{vahidnia2025multi,
  title={{Multi-Agent} systems of large language models as weight assigners: An approach to collaborative weighting in spatial multi-criteria decision-making},
  author={Vahidnia, Mohammad H},
  journal={Geomatica},
  pages={100071},
  year={2025},
  publisher={Elsevier}
}

@article{luo2025geojson,
  title={{GeoJSON} Agents: A Multi-Agent {LLM} Architecture for Geospatial Analysis-Function Calling vs Code Generation},
  author={Luo, Qianqian and Xu, Liuchang and Lin, Qingming and Wu, Sensen and Mao, Ruichen and Wang, Chao and Feng, Hailin and Huang, Bo and Du, Zhenhong},
  journal={arXiv preprint arXiv:2509.08863},
  year={2025}
}

@article{wang2025cartoagent,
  title={{CartoAgent}: a multimodal large language model-powered multi-agent cartographic framework for map style transfer and evaluation},
  author={Wang, Chenglong and Kang, Yuhao and Gong, Zhaoya and Zhao, Pengjun and Feng, Yu and Zhang, Wenjia and Li, Ge},
  journal={International Journal of Geographical Information Science},
  pages={1--34},
  year={2025},
  publisher={Taylor \& Francis}
}

@inproceedings{tsokanaridou2025da4dte,
  title={{DA4DTE}: An agentic system for enhancing the accessibility of Digital Twins of Earth},
  author={Tsokanaridou, M and Hackstein, J and Hoxha, G and Kefalidis, SA and Plas, K and Demir, B and Koubarakis, M and Corsi, M and Leoni, C and Pasquali, G and others},
booktitle={Workshop on AI-driven Data Engineering and Reusability for Earth and Space Sciences},
  year={2025}
}

@inproceedings{guo2024remote,
  title={Remote sensing chatgpt: Solving remote sensing tasks with chatgpt and visual models},
  author={Guo, Haonan and Su, Xin and Wu, Chen and Du, Bo and Zhang, Liangpei and Li, Deren},
  booktitle={IGARSS 2024-2024 IEEE International Geoscience and Remote Sensing Symposium},
  pages={11474--11478},
  year={2024},
  organization={IEEE}
}

@article{xiao2025llm,
  title={{LLM} agent framework for intelligent change analysis in urban environment using remote sensing imagery},
  author={Xiao, Zixuan and Ma, Jun},
  journal={Automation in Construction},
  volume={177},
  pages={106341},
  year={2025},
  publisher={Elsevier}
}

@inproceedings{chu2024geoagent,
  title={Automating Geospatial Vision Tasks with a Large Language Model Agent},
  author={Chen, Yuxing and Wang, Weijie and Kurtz, Camille and Lobry, Sylvain},
  booktitle={Joint European Conference on Machine Learning and Knowledge Discovery in Databases},
  pages={218--235},
  year={2025},
  organization={Springer}
}

@article{liu2024remoteclip,
  title={Remoteclip: A vision language foundation model for remote sensing},
  author={Liu, Fan and Chen, Delong and Guan, Zhangqingyun and Zhou, Xiaocong and Zhu, Jiale and Ye, Qiaolin and Fu, Liyong and Zhou, Jun},
  journal={IEEE Transactions on Geoscience and Remote Sensing},
  volume={62},
  pages={1--16},
  year={2024},
  publisher={IEEE}
}

@article{xu2024rs,
  title={{RS-Agent}: Automating Remote Sensing Tasks through Intelligent Agent},
  author={Xu, Wenjia and Yu, Zijian and Mu, Boyang and Wei, Zhiwei and Zhang, Yuanben and Li, Guangzuo and Peng, Mugen},
  journal={arXiv preprint arXiv:2406.07089},
  year={2024}
}

@article{guo2025deepseek,
  title={{DeepSeek-R1}: Incentivizing reasoning capability in llms via reinforcement learning},
  author={Guo, Daya and Yang, Dejian and Zhang, Haowei and Song, Junxiao and Zhang, Ruoyu and Xu, Runxin and Zhu, Qihao and Ma, Shirong and Wang, Peiyi and Bi, Xiao and others},
  journal={arXiv preprint arXiv:2501.12948},
  year={2025}
}

@article{hu2025ringmo,
  title={{RINGMO-Agent}: A Unified Remote Sensing Foundation Model for Multi-Platform and Multi-Modal Reasoning},
  author={Hu, Huiyang and Wang, Peijin and Feng, Yingchao and Wei, Kaiwen and Yin, Wenxin and Diao, Wenhui and Wang, Mengyu and Bi, Hanbo and Kang, Kaiyue and Ling, Tong and others},
  journal={arXiv preprint arXiv:2507.20776},
  year={2025}
}

@article{bhattaram2025geoflow,
  title={{GeoFlow}: Agentic Workflow Automation for Geospatial Tasks},
  author={Bhattaram, Amulya and Chung, Justin and Chung, Stanley and Gupta, Ranit and Ramamoorthy, Janani and Gullapalli, Kartikeya and Marculescu, Diana and Stamoulis, Dimitrios},
  journal={arXiv preprint arXiv:2508.04719},
  year={2025}
}

@article{shabbir2025thinkgeo,
  title={{THINKGEO}: Evaluating Tool-Augmented Agents for Remote Sensing Tasks},
  author={Shabbir, Akashah and Munir, Muhammad Akhtar and Dudhane, Akshay and Sheikh, Muhammad Umer and Khan, Muhammad Haris and Fraccaro, Paolo and Moreno, Juan Bernabe and Khan, Fahad Shahbaz and Khan, Salman},
  journal={arXiv preprint arXiv:2505.23752},
  year={2025}
}

@article{zhang2025geospatial,
  title={Geospatial large language model trained with a simulated environment for generating tool-use chains autonomously},
  author={Zhang, Yifan and Li, Jingxuan and Wang, Zhiyun and He, Zhengting and Guan, Qingfeng and Lin, Jianfeng and Yu, Wenhao},
  journal={International Journal of Applied Earth Observation and Geoinformation},
  volume={136},
  pages={104312},
  year={2025},
  publisher={Elsevier}
}

@article{guo2025earthlink,
  title={{EarthLink}: A Self-Evolving AI Agent for Climate Science},
  author={Guo, Zijie and Wang, Jiong and Yue, Xiaoyu and Wei, Wangxu and Jiang, Zhe and Xu, Wanghan and Fei, Ben and Zhang, Wenlong and Gu, Xinyu and Cheng, Lijing and others},
  journal={arXiv preprint arXiv:2507.17311},
  year={2025}
}

@article{gupta2024geode,
  title={Geode: A Zero-shot Geospatial Question-Answering Agent with Explicit Reasoning and Precise Spatio-Temporal Retrieval},
  author={Gupta, Devashish Vikas and Ishaqui, Azeez Syed Ali and Kadiyala, Divya Kiran},
  journal={arXiv preprint arXiv:2407.11014},
  year={2024}
}

@article{akinboyewa2025gis,
  title={{GIS} copilot: Towards an autonomous {GIS} agent for spatial analysis},
  author={Akinboyewa, Temitope and Li, Zhenlong and Ning, Huan and Lessani, M Naser},
  journal={International Journal of Digital Earth},
  volume={18},
  number={1},
  pages={2497489},
  year={2025},
  publisher={Taylor \& Francis}
}

@inproceedings{geoagent,
  title     = {GeoAgent: To Empower LLMs Using Geospatial Tools for Address Standardization},
  author    = {Huang, Cheng and Zhang, Yifan and Wang, Zhiyun and Yu, Wenhao},
  booktitle = {Findings of the Association for Computational Linguistics ACL},
  year      = {2024}
}

@article{earthbench,
  title   = {Earth-Agent: Unlocking the Full Landscape of Earth Observation with Agents},
  author  = {Feng, Peilin and Lv, Zhutao and Ye, Junyan and Wang, Xiaolei and Huo, Xinjie and Yu, Jinhua and Xu, Wanghan and Zhang, Wenlong and Bai, Lei and He, Conghui and Li, Weijia},
  journal = {arXiv preprint arXiv:2509.23141},
  year    = {2025}
}

@inproceedings{1krechetova2025geobenchx,
author = {Krechetova, Varvara and Kochedykov, Denis},
title = {{GeoBenchX}: Benchmarking {LLMs} in Agent Solving Multistep Geospatial Tasks},
year = {2025},
isbn = {9798400722615},
publisher = {Association for Computing Machinery},
address = {New York, NY, USA},
url = {https://doi.org/10.1145/3764915.3770721},
doi = {10.1145/3764915.3770721},
booktitle = {Proceedings of the 1st ACM SIGSPATIAL International Workshop on Generative and Agentic AI for Multi-Modality Space-Time Intelligence},
pages = {27–35},
numpages = {9}
}

@article{hou2025geocode,
  title={GeoCode-GPT: A large language model for geospatial code generation},
  author={Hou, Shuyang and Shen, Zhangxiao and Zhao, Anqi and Liang, Jianyuan and Gui, Zhipeng and Guan, Xuefeng and Li, Rui and Wu, Huayi},
  journal={International Journal of Applied Earth Observation and Geoinformation},
  pages={104456},
  year={2025},
  publisher={Elsevier}
}

@inproceedings{toolemu,
  title     = {Identifying the Risks of LM Agents with an LM-Emulated Sandbox},
  author    = {Ruan, Yangjun and Dong, Honghua and Wang, Andrew and Pitis, Silviu and Zhou, Yongchao and Ba, Jimmy and Dubois, Yann and Maddison, Chris J. and Hashimoto, Tatsunori},
  booktitle = {International Conference on Learning Representations (ICLR)},
  year      = {2024}
}

@article{geographrag,
  title={{GeoGraphRAG}: A graph-based retrieval-augmented generation approach for empowering large language models in automated geospatial modeling},
  author={Liang, Jianyuan and Hou, Shuyang and Jiao, Haoyue and Qing, Yaxian and Zhao, Anqi and Shen, Zhangxiao and Xiang, Longgang and Wu, Huayi},
  journal={International Journal of Applied Earth Observation and Geoinformation},
  volume={142},
  pages={104712},
  year={2025},
  publisher={Elsevier}
}

@misc{earthai,
  title        = {Google Earth AI and Gemini for Climate and Environmental Analysis},
  author       = {{Google Earth Team}},
  howpublished = {\url{https://earth.google.com}},
  note         = {Accessed 2025},
  year         = {2025}
}

@article{bommasani2021opportunities,
  title   = {On the Opportunities and Risks of Foundation Models},
  author  = {Bommasani, Rishi and Hudson, Drew A. and Adeli, Ehsan and Altman, Russ and Arora, Simran and von Arx, Sydney and Bernstein, Michael S. and Bohg, Jeannette and Bosselut, Antoine and Brunskill, Emma and Brynjolfsson, Erik and Buch, Shyamal and Card, Dallas and Castellon, Rodrigo and Chatterji, Niladri and Chen, Annie S. and Creel, Kathleen and Davis, Jared Quincy and Demszky, Dora and Donahue, Chris and others},
  journal = {arXiv preprint arXiv:2108.07258},
  year    = {2021}
}

@inproceedings{krizhevsky2012imagenet,
  title     = {ImageNet Classification with Deep Convolutional Neural Networks},
  author    = {Krizhevsky, Alex and Sutskever, Ilya and Hinton, Geoffrey E.},
  booktitle = {Advances in Neural Information Processing Systems},
  volume    = {25},
  year      = {2012}
}

@article{simonyan2014very,
  title={Very deep convolutional networks for large-scale image recognition},
  author={Simonyan, Karen and Zisserman, Andrew},
  journal={arXiv preprint arXiv:1409.1556},
  year={2014}
}

@inproceedings{he2016deep,
  title={Deep residual learning for image recognition},
  author={He, Kaiming and Zhang, Xiangyu and Ren, Shaoqing and Sun, Jian},
  booktitle={Proceedings of the IEEE conference on computer vision and pattern recognition},
  pages={770--778},
  year={2016}
}

@inproceedings{xie2022simmim,
  title     = {{SimMIM}: A Simple Framework for Masked Image Modeling},
  author    = {Xie, Zhenda and Zhang, Zheng and Cao, Yue and Lin, Yutong and Bao, Jianmin and Yao, Zhuliang and Dai, Qi and Hu, Han},
  booktitle = {Proceedings of the IEEE/CVF Conference on Computer Vision and Pattern Recognition (CVPR)},
  pages     = {9643--9653},
  year      = {2022}
}

@article{guo2025agentsense,
  title={{AgentSense}: {LLMs} Empower Generalizable and Explainable Web-Based Participatory Urban Sensing},
  author={Guo, Xusen and Peng, Mingxing and Hao, Xixuan and Zou, Xingchen and Wang, Qiongyan and Ruan, Sijie and Liang, Yuxuan},
  journal={arXiv preprint arXiv:2510.19661},
  year={2025}
}

@inproceedings{yang2010bag,
  title={Bag-of-visual-words and spatial extensions for land-use classification},
  author={Yang, Yi and Newsam, Shawn},
  booktitle={Proceedings of the 18th SIGSPATIAL international conference on advances in geographic information systems},
  pages={270--279},
  year={2010}
}

@article{xia2017aid,
  title={AID: A benchmark data set for performance evaluation of aerial scene classification},
  author={Xia, Gui-Song and Hu, Jingwen and Hu, Fan and Shi, Baoguang and Bai, Xiang and Zhong, Yanfei and Zhang, Liangpei and Lu, Xiaoqiang},
  journal={IEEE Transactions on Geoscience and Remote Sensing},
  volume={55},
  number={7},
  pages={3965--3981},
  year={2017},
  publisher={IEEE}
}

@article{cheng2017remote,
  title={Remote sensing image scene classification: Benchmark and state of the art},
  author={Cheng, Gong and Han, Junwei and Lu, Xiaoqiang},
  journal={Proceedings of the IEEE},
  volume={105},
  number={10},
  pages={1865--1883},
  year={2017},
  publisher={IEEE}
}

@article{li2025lhrs,
  title={{LHRS-Bot-Nova}: Improved multimodal large language model for remote sensing vision-language interpretation},
  author={Li, Zhenshi and Muhtar, Dilxat and Gu, Feng and He, Yanglangxing and Zhang, Xueliang and Xiao, Pengfeng and He, Guangjun and Zhu, Xiaoxiang},
  journal={ISPRS Journal of Photogrammetry and Remote Sensing},
  volume={227},
  pages={539--550},
  year={2025},
  publisher={Elsevier}
}

@article{zhang2025georsmllm,
  title={{GeoRSMLLM}: A Multimodal Large Language Model for Vision-Language Tasks in Geoscience and Remote Sensing},
  author={Zhang, Zilun and Shen, Haozhan and Zhao, Tiancheng and Chen, Bin and Guan, Zian and Wang, Yuhao and Jia, Xu and Cai, Yuxiang and Shang, Yongheng and Yin, Jianwei},
  journal={arXiv preprint arXiv:2503.12490},
  year={2025}
}

@article{yuan2025omnigeo,
  title={{OmniGeo}: Towards a multimodal large language models for geospatial artificial intelligence},
  author={Yuan, Long and Mo, Fengran and Huang, Kaiyu and Wang, Wenjie and Zhai, Wangyuxuan and Zhu, Xiaoyu and Li, You and Xu, Jinan and Nie, Jian-Yun},
  journal={arXiv preprint arXiv:2503.16326},
  year={2025}
}

@article{wei2025geotool,
  title={{GeoTool-GPT}: a trainable method for facilitating Large Language Models to master GIS tools},
  author={Wei, Cheng and Zhang, Yifan and Zhao, Xinru and Zeng, Ziyi and Wang, Zhiyun and Lin, Jianfeng and Guan, Qingfeng and Yu, Wenhao},
  journal={International Journal of Geographical Information Science},
  volume={39},
  number={4},
  pages={707--731},
  year={2025},
  publisher={Taylor \& Francis}
}

@inproceedings{xu2025agentic,
  title={Agentic {LLM} Framework for Generating Spatial Intelligence to Support Decision-Making in Smart Cities},
  author={Xu, Yufei and Kibria, Gulam and Peeta, Srinivas},
  booktitle={Proceedings of the 1st ACM SIGSPATIAL International Workshop on Spatial Intelligence for Smart and Connected Communities},
  pages={63--71},
  year={2025}
}

@article{helber2019eurosat,
  title={Eurosat: A novel dataset and deep learning benchmark for land use and land cover classification},
  author={Helber, Patrick and Bischke, Benjamin and Dengel, Andreas and Borth, Damian},
  journal={IEEE Journal of Selected Topics in Applied Earth Observations and Remote Sensing},
  volume={12},
  number={7},
  pages={2217--2226},
  year={2019},
  publisher={IEEE}
}

@article{long2021creating,
  title={On creating benchmark dataset for aerial image interpretation: Reviews, guidances, and million-aid},
  author={Long, Yang and Xia, Gui-Song and Li, Shengyang and Yang, Wen and Yang, Michael Ying and Zhu, Xiao Xiang and Zhang, Liangpei and Li, Deren},
  journal={IEEE Journal of selected topics in applied earth observations and remote sensing},
  volume={14},
  pages={4205--4230},
  year={2021},
  publisher={IEEE}
}

@article{qi2020mlrsnet,
  title={MLRSNet: A multi-label high spatial resolution remote sensing dataset for semantic scene understanding},
  author={Qi, Xiaoman and Zhu, Panpan and Wang, Yuebin and Zhang, Liqiang and Peng, Junhuan and Wu, Mengfan and Chen, Jialong and Zhao, Xudong and Zang, Ning and Mathiopoulos, P Takis},
  journal={ISPRS Journal of Photogrammetry and Remote Sensing},
  volume={169},
  pages={337--350},
  year={2020},
  publisher={Elsevier}
}

@inproceedings{maggiori2017can,
  title={Can semantic labeling methods generalize to any city? the inria aerial image labeling benchmark},
  author={Maggiori, Emmanuel and Tarabalka, Yuliya and Charpiat, Guillaume and Alliez, Pierre},
  booktitle={2017 IEEE International geoscience and remote sensing symposium (IGARSS)},
  pages={3226--3229},
  year={2017},
  organization={IEEE}
}

@article{wang2021loveda,
  title={LoveDA: A remote sensing land-cover dataset for domain adaptive semantic segmentation},
  author={Wang, Junjue and Zheng, Zhuo and Ma, Ailong and Lu, Xiaoyan and Zhong, Yanfei},
  journal={arXiv preprint arXiv:2110.08733},
  year={2021}
}

@inproceedings{toker2022dynamicearthnet,
  title={Dynamicearthnet: Daily multi-spectral satellite dataset for semantic change segmentation},
  author={Toker, Aysim and Kondmann, Lukas and Weber, Mark and Eisenberger, Marvin and Camero, Andr{\'e}s and Hu, Jingliang and Hoderlein, Ariadna Pregel and {\c{S}}enaras, {\c{C}}a{\u{g}}lar and Davis, Timothy and Cremers, Daniel and others},
  booktitle={Proceedings of the IEEE/CVF Conference on Computer Vision and Pattern Recognition},
  pages={21158--21167},
  year={2022}
}

@article{brown2022dynamic,
  title={Dynamic World, Near real-time global 10 m land use land cover mapping},
  author={Brown, Christopher F and Brumby, Steven P and Guzder-Williams, Brookie and Birch, Tanya and Hyde, Samantha Brooks and Mazzariello, Joseph and Czerwinski, Wanda and Pasquarella, Valerie J and Haertel, Robert and Ilyushchenko, Simon and others},
  journal={Scientific data},
  volume={9},
  number={1},
  pages={251},
  year={2022},
  publisher={Nature Publishing Group UK London}
}

@inproceedings{xia2018dota,
  title={DOTA: A large-scale dataset for object detection in aerial images},
  author={Xia, Gui-Song and Bai, Xiang and Ding, Jian and Zhu, Zhen and Belongie, Serge and Luo, Jiebo and Datcu, Mihai and Pelillo, Marcello and Zhang, Liangpei},
  booktitle={Proceedings of the IEEE conference on computer vision and pattern recognition},
  pages={3974--3983},
  year={2018}
}

@article{lam2018xview,
  title={xview: Objects in context in overhead imagery},
  author={Lam, Darius and Kuzma, Richard and McGee, Kevin and Dooley, Samuel and Laielli, Michael and Klaric, Matthew and Bulatov, Yaroslav and McCord, Brendan},
  journal={arXiv preprint arXiv:1802.07856},
  year={2018}
}

@article{sun2022fair1m,
  title={FAIR1M: A benchmark dataset for fine-grained object recognition in high-resolution remote sensing imagery},
  author={Sun, Xian and Wang, Peijin and Yan, Zhiyuan and Xu, Feng and Wang, Ruiping and Diao, Wenhui and Chen, Jin and Li, Jihao and Feng, Yingchao and Xu, Tao and others},
  journal={ISPRS Journal of Photogrammetry and Remote Sensing},
  volume={184},
  pages={116--130},
  year={2022},
  publisher={Elsevier}
}

@article{CORE,
  title={{CORE}: Full-Path Evaluation of {LLM} Agents Beyond Final State},
  author={Zuo, Yutong and Wang, Zirui and Zhang, Jiaxin and Wu, Yilun and Li, Bo and Zhu, Erpeng and Jiang, Lihong and Zhang, Xifeng and Yau, Stanley K. S. and Lin, Zhaoyuan and others},
  journal={arXiv preprint arXiv:2407.03728}, 
  year={2024},
}

@article{chen2020spatial,
  title={A spatial-temporal attention-based method and a new dataset for remote sensing image change detection},
  author={Chen, Hao and Shi, Zhenwei},
  journal={Remote sensing},
  volume={12},
  number={10},
  pages={1662},
  year={2020},
  publisher={MDPI}
}

@article{shi2021deeply,
  title={A deeply supervised attention metric-based network and an open aerial image dataset for remote sensing change detection},
  author={Shi, Qian and Liu, Mengxi and Li, Shengchen and Liu, Xiaoping and Wang, Fei and Zhang, Liangpei},
  journal={IEEE transactions on geoscience and remote sensing},
  volume={60},
  pages={1--16},
  year={2021},
  publisher={IEEE}
}

@article{shen2021s2looking,
  title={S2Looking: A satellite side-looking dataset for building change detection},
  author={Shen, Li and Lu, Yao and Chen, Hao and Wei, Hao and Xie, Donghai and Yue, Jiabao and Chen, Rui and Lv, Shouye and Jiang, Bitao},
  journal={Remote Sensing},
  volume={13},
  number={24},
  pages={5094},
  year={2021},
  publisher={MDPI}
}

@inproceedings{demir2018deepglobe,
  title={Deepglobe 2018: A challenge to parse the earth through satellite images},
  author={Demir, Ilke and Koperski, Krzysztof and Lindenbaum, David and Pang, Guan and Huang, Jing and Basu, Saikat and Hughes, Forest and Tuia, Devis and Raskar, Ramesh},
  booktitle={Proceedings of the IEEE conference on computer vision and pattern recognition workshops},
  pages={172--181},
  year={2018}
}

@InProceedings{DeepGlobe18,
 author = {Demir, Ilke and Koperski, Krzysztof and Lindenbaum, David and Pang, Guan and Huang, Jing and Basu, Saikat and Hughes, Forest and Tuia, Devis and Raskar, Ramesh},
 title = {DeepGlobe 2018: A Challenge to Parse the Earth Through Satellite Images},
 booktitle = {The IEEE Conference on Computer Vision and Pattern Recognition (CVPR) Workshops},
 month = {June},
 year = {2018}
}

@article{adimoolam2023efficient,
  title={Efficient deduplication and leakage detection in large scale image datasets with a focus on the crowdai mapping challenge dataset},
  author={Adimoolam, Yeshwanth Kumar and Chatterjee, Bodhiswatta and Poullis, Charalambos and Averkiou, Melinos},
  journal={arXiv preprint arXiv:2304.02296},
  year={2023}
}

@article{gupta2019xbd,
  title={xbd: A dataset for assessing building damage from satellite imagery},
  author={Gupta, Ritwik and Hosfelt, Richard and Sajeev, Sandra and Patel, Nirav and Goodman, Bryce and Doshi, Jigar and Heim, Eric and Choset, Howie and Gaston, Matthew},
  journal={arXiv preprint arXiv:1911.09296},
  year={2019}
}

@article{rahnemoonfar2021floodnet,
  title={Floodnet: A high resolution aerial imagery dataset for post flood scene understanding},
  author={Rahnemoonfar, Maryam and Chowdhury, Tashnim and Sarkar, Argho and Varshney, Debvrat and Yari, Masoud and Murphy, Robin Roberson},
  journal={IEEE Access},
  volume={9},
  pages={89644--89654},
  year={2021},
  publisher={IEEE}
}

@inproceedings{bonafilia2020sen1floods11,
  title={Sen1Floods11: A georeferenced dataset to train and test deep learning flood algorithms for sentinel-1},
  author={Bonafilia, Derrick and Tellman, Beth and Anderson, Tyler and Issenberg, Erica},
  booktitle={Proceedings of the IEEE/CVF conference on computer vision and pattern recognition workshops},
  pages={210--211},
  year={2020}
}

@inproceedings{shen2023firerisk,
  title={Firerisk: A remote sensing dataset for fire risk assessment with benchmarks using supervised and self-supervised learning},
  author={Shen, Shuchang and Seneviratne, Sachith and Wanyan, Xinye and Kirley, Michael},
  booktitle={2023 international conference on digital image computing: techniques and applications (DICTA)},
  pages={189--196},
  year={2023},
  organization={IEEE}
}

@inproceedings{zhao2024urbansarfloods,
  title={UrbanSARFloods: Sentinel-1 SLC-based benchmark dataset for urban and open-area flood mapping},
  author={Zhao, Jie and Xiong, Zhitong and Zhu, Xiao Xiang},
  booktitle={Proceedings of the IEEE/CVF Conference on Computer Vision and Pattern Recognition},
  pages={419--429},
  year={2024}
}

@inproceedings{yamani2024remote,
  title={Remote Sensing Image Captioning Using Deep Learning},
  author={Yamani, Bhavitha and Medavarapu, Nikhil and Rakesh, S},
  booktitle={2024 International Conference on Automation and Computation (AUTOCOM)},
  pages={295--302},
  year={2024},
  organization={IEEE}
}

@article{zheng2021mutual,
  title={Mutual attention inception network for remote sensing visual question answering},
  author={Zheng, Xiangtao and Wang, Binqiang and Du, Xingqian and Lu, Xiaoqiang},
  journal={IEEE Transactions on Geoscience and Remote Sensing},
  volume={60},
  pages={1--14},
  year={2021},
  publisher={IEEE}
}

@article{wang2023ssl4eo,
  title={SSL4EO-S12: A large-scale multimodal, multitemporal dataset for self-supervised learning in Earth observation [Software and Data Sets]},
  author={Wang, Yi and Braham, Nassim Ait Ali and Xiong, Zhitong and Liu, Chenying and Albrecht, Conrad M and Zhu, Xiao Xiang},
  journal={IEEE Geoscience and Remote Sensing Magazine},
  volume={11},
  number={3},
  pages={98--106},
  year={2023},
  publisher={IEEE}
}

@inproceedings{velazquez2025earthview,
  title={EarthView: a large scale remote sensing dataset for self-supervision},
  author={Velazquez, Diego and Rodriguez, Pau and Alonso, Sergio and Gonfaus, Josep M and Gonzalez, Jordi and Richarte, Gerardo and Marin, Javier and Bengio, Yoshua and Lacoste, Alexandre},
  booktitle={Proceedings of the Winter Conference on Applications of Computer Vision},
  pages={1228--1237},
  year={2025}
}

@inproceedings{alipour2025disa,
  title={DiSa: Directional Saliency-Aware Prompt Learning for Generalizable Vision-Language Models},
  author={Alipour Talemi, Niloufar and Kashiani, Hossein and Nowdeh, Hossein R and Afghah, Fatemeh},
  booktitle={Proceedings of the 31st ACM SIGKDD Conference on Knowledge Discovery and Data Mining V. 2},
  pages={37--46},
  year={2025}
}

@inproceedings{talemi2025style,
  title={Style-Pro: Style-Guided Prompt Learning for Generalizable Vision-Language Models},
  author={Talemi, Niloufar Alipour and Kashiani, Hossein and Afghah, Fatemeh},
  booktitle={2025 IEEE/CVF Winter Conference on Applications of Computer Vision (WACV)},
  pages={6207--6216},
  year={2025},
  organization={IEEE}
}

@inproceedings{kashiani2025roads,
  title={Roads: Robust prompt-driven multi-class anomaly detection under domain shift},
  author={Kashiani, Hossein and Talemi, Niloufar Alipour and Afghah, Fatemeh},
  booktitle={2025 IEEE/CVF Winter Conference on Applications of Computer Vision (WACV)},
  pages={7908--7917},
  year={2025},
  organization={IEEE}
}

@article{wang2025non,
  title={Non-local and local feature-coupled self-supervised network for hyperspectral anomaly detection},
  author={Wang, Degang and Ren, Longfei and Sun, Xu and Gao, Lianru and Chanussot, Jocelyn},
  journal={IEEE Journal of Selected Topics in Applied Earth Observations and Remote Sensing},
  year={2025},
  publisher={IEEE}
}
}

\end{document}